\documentclass[10pt, a4paper]{article}
 \usepackage{multirow} 
 \usepackage{booktabs}
 \usepackage{url}
\usepackage[dvipsnames]{xcolor}
\usepackage{makecell}

\usepackage[]{lrec-coling2024}
\title{
Small LLMs for Medical NLP: a Systematic Analysis of Few-Shot, Constraint Decoding, Fine-Tuning and Continual Pre-Training in Italian 
}

\name{Pietro Ferrazzi\textsuperscript{1,2}, Mattia Franzin\textsuperscript{1}, Alberto Lavelli\textsuperscript{1}, Bernardo Magnini\textsuperscript{1}} 

\address{\textsuperscript{1}Fondazione Bruno Kessler, Povo, Trento, Italy \\ \textsuperscript{2}University of Padova, Padova, Italy}

\abstract{
Large Language Models (LLMs) consistently excel in diverse medical Natural Language Processing (NLP) tasks, yet their substantial computational requirements often limit deployment in real-world healthcare settings. 
In this work, we investigate whether "small" LLMs (around one billion parameters) can effectively perform medical tasks while maintaining competitive accuracy. 
We evaluate models from three major families—Llama-3, Gemma-3, and Qwen3—across 20 clinical NLP tasks among Named Entity Recognition, Relation Extraction, Case Report Form Filling, Question Answering, and Argument Mining.
We systematically compare a range of adaptation strategies, both at inference time (few-shot prompting, constraint decoding) and at training time (supervised fine-tuning, continual pretraining). 
Fine-tuning emerges as the most effective approach, while the combination of few-shot prompting and constraint decoding offers strong lower-resource alternatives. 
Our results show that small LLMs can match or even surpass larger baselines, with our best configuration based on Qwen3-1.7B achieving an average score +9.2 points higher than Qwen3-32B. 
We release a comprehensive collection of all the publicly available Italian medical datasets for NLP tasks, together with our top-performing models.
Furthermore, we release an Italian dataset of 126M words from the Emergency Department of an Italian Hospital, and 175M words from various sources that we used for continual pre-training.
 \\ \newline  }

\begin{document}

\maketitleabstract

\section{Introduction}

\begin{figure*}[t]
  \includegraphics[width=0.48\linewidth]{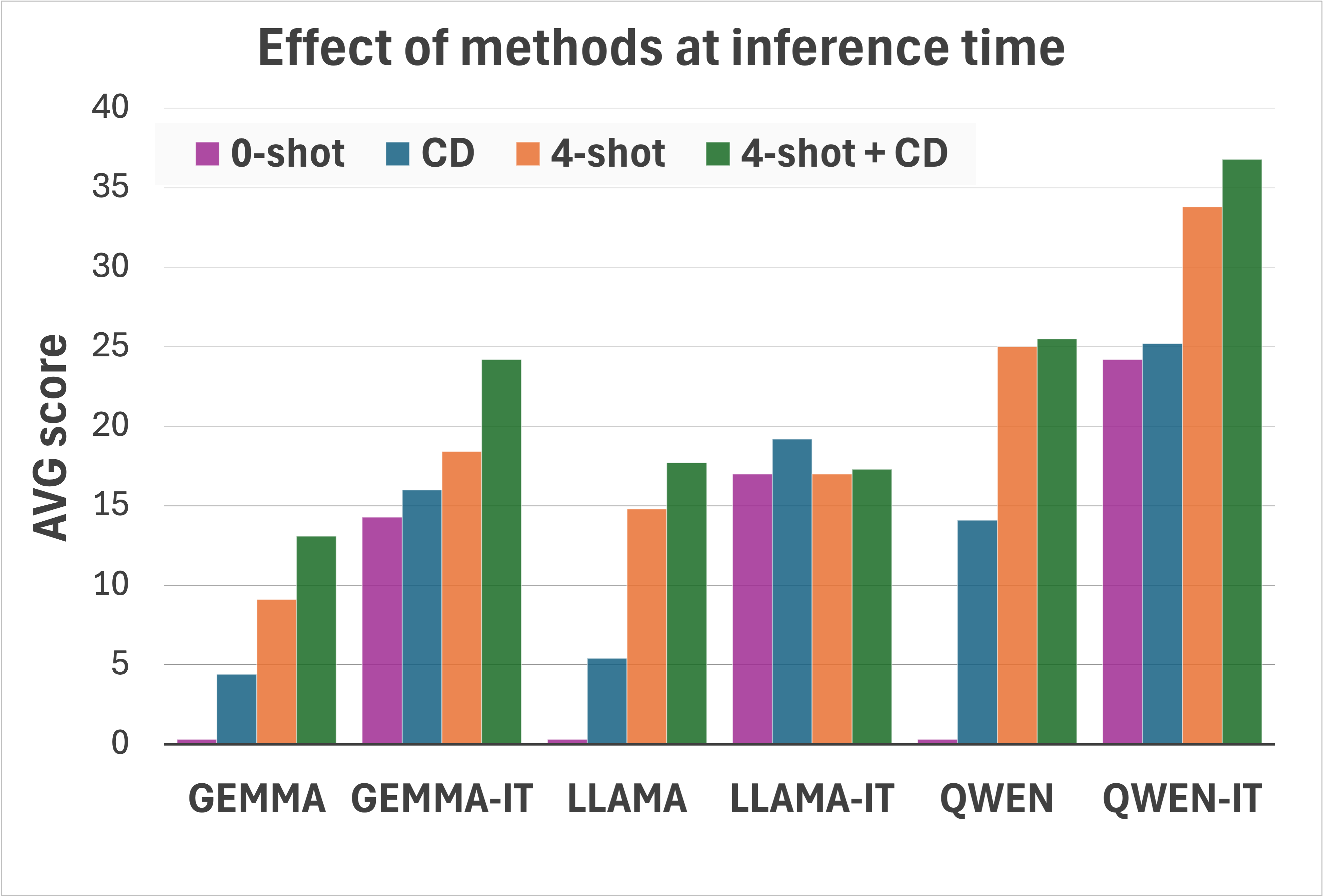} \hfill
  \includegraphics[width=0.48\linewidth]{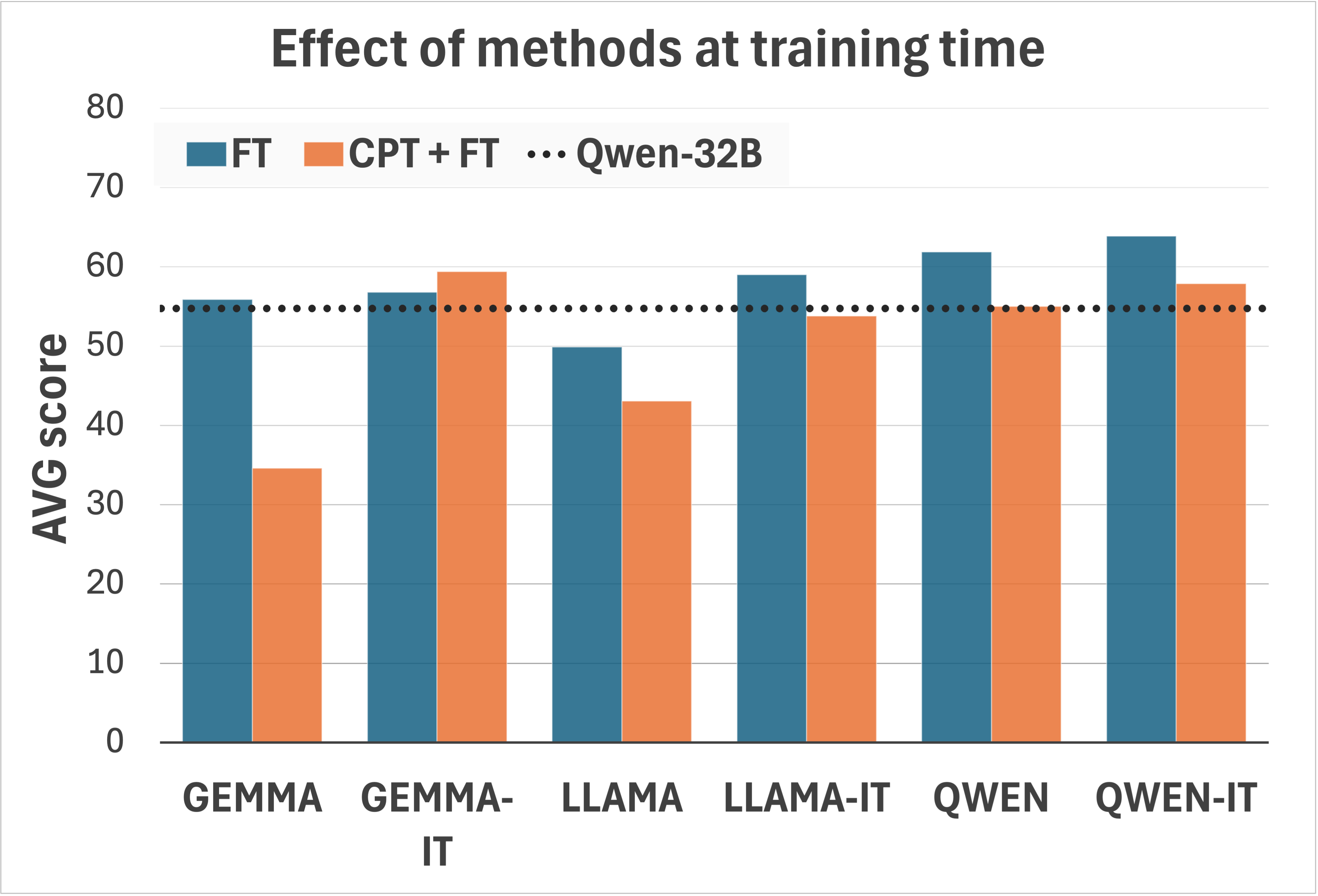} \hfill
  \caption {Average performances of 1B LLMs on the 14 medical sub-tasks when different methods are applied at both inference (\textbf{left}) and training (\textbf{right}) time. Exposing models to Fine-Tuning (\textit{FT}) turns out to be the most effective approach overall, consistently outperforming the baseline (Qwen3-32B with 4-shot). 
  Continual Pre-Training (\textit{CPT}) has a positive impact with respect to simple FT only in one case (gemma-3-1b-it).
  \textit{4-shot} is consistently better than Constraint Decoding (\textit{CD}), and the combination of the two shows to be beneficial.  }
  \label{fig:impact}
\end{figure*}

Large Language Models (LLMs) have achieved remarkable performance across a wide range of medical Natural Language Processing (NLP) tasks, from clinical concept extraction to question answering. 
Recently, attention has turned to the so-called "small" LLMs (SLLMs)—models with around one billion parameters—which have become the focus of intensive research \citep{10.1145/3711896.3736563}. 
This shift raises a new question: can small, resource-efficient LLMs perform medical tasks effectively? The answer carries significant practical implications, as hospitals, clinics, and healthcare organisations often operate under strict computational and financial constraints that make large-scale models impractical to deploy \cite{10651607}. 
To this extent, a plethora of methods has emerged, 
aiming to enhance effectiveness and adapt SLLMs to specialized objectives and downstream applications, including few-shot prompting, constraint decoding, fine-tuning and continual pre-training.
Such approaches attempt to overtake the boundaries that SLLMs' parametric knowledge often exhibits with respect to larger models and the challenges of providing consistent, structured outputs.\\
In this work, we investigate the state of SLLMs in a range of medical NLP  tasks in Italian, systematically assessing the impact of different techniques and identifying the most effective methods.
We build our analysis on a curated and comprehensive collection of all publicly available Italian datasets for medical NLP tasks. By systematically evaluating different adaptation strategies, we identify the most effective approaches and ultimately obtain a single compact model that outperforms models up to 30 times larger. To enable the evaluation of continual pre-training, we collect  a large dataset of medical text from several sources.
Our contribution can be summarized as follows:
\begin{itemize}
    \item We provide a systematic evaluation of the impact of few-shot prompting, constraint decoding, fine-tuning, and continual pre-training on SLLMs for Italian medical tasks.
\end{itemize}
To reach the evaluation goals, we curate and release the following resources\footnote{\url{huggingface.co/collections/NLP-FBK/small-llms-for-medical-tasks-italian}}:
\begin{itemize}
    \item the first comprehensive collection of datasets for NLP medical tasks in Italian;
    \item a new medical dataset in Italian composed of 300 million words from both clinical settings and various sources;
    \item a Small LLM that outperforms bigger models on medical tasks in Italian.
\end{itemize}
We designed our publicly available codebase\footnote{\url{github.com/ferrazzipietro/llms-for-medical-nlp}} to be easily extensible to new tasks and models, hoping to provide a useful tool to researchers.

\section{Methods}\label{sec:methods}

In this work, we evaluate how effectively small LLMs can perform on a diverse set of medical tasks, exploring how different adaptation strategies can push their performance to the limit. Our goal is to develop a single, versatile model capable of handling multiple clinical tasks. 
To achieve such a result, we adopt several strategies that have been proposed to specialise LLMs for domain-specific applications, both at inference and training time. 
Zero-Shot approaches rely only on models' pre-trained knowledge, while Constraint Decoding \citep{geng2024grammarconstraineddecodingstructurednlp} and Few-Shot learning \cite{DBLP:conf/nips/BrownMRSKDNSSAA20} focus on enforcing pre-defined output structures.
Supervised Fine-Tuning aligns the model with complex instructions \cite{DBLP:journals/corr/abs-2308-10792}, while continual pretraining exposes the model to domain-specific data to improve its medical knowledge \citep{DBLP:journals/csur/ShiXWQWWWEW26}. 

\paragraph{Zero-shot prompting} We first assess model performance in a zero-shot setting, where the model is only provided with a description of the task and the input text on which to perform it. In this scenario, the model relies entirely on its pre-trained knowledge to generate outputs, without seeing any task-specific examples. Zero-shot evaluation serves as a baseline for understanding how well the model can generalize to medical NLP tasks out of the box.

\paragraph{Constraint decoding} Generating structured outputs is critical for real-world systems, which often require formatted outputs to be automatically integrated into downstream components. 
Correct predictions might turn unusable when presenting formatting errors.
Therefore, we employ a constraint decoding method based on \textit{outlines}\footnote{\url{https://github.com/dottxt-ai/outlines}}. 
Once a valid output schema is defined (e.g., a JSON template), the models' decoding process dynamically masks out all tokens that would lead to an invalid structure at each generation step. For instance, if the model is required to generate a valid JSON string, the decoder ensures that brackets and quotation marks are properly closed and that keys and values follow the expected syntax.
This approach restricts the model’s generation to comply with the expected format, reducing parsing errors and ensuring consistency \citep{geng2024grammarconstraineddecodingstructurednlp}.

\paragraph{Few-shots} Another widely used approach to enforce output formatting is few-shot prompting \cite{DBLP:conf/nips/BrownMRSKDNSSAA20}, where the model is provided with a small number of examples demonstrating the task. These examples help the model better understand the task instructions, improve adherence to the expected output format, and often lead to substantial performance gains, especially for smaller models that may not generalize as well as bigger ones from pre-training alone. We select the examples by random sampling from the training split of each dataset.

\paragraph{Instruction-tuning} Fine-tuning the model on a dataset of task instructions paired with expected outputs is an effective strategy to enhance LLM per \cite{DBLP:journals/corr/abs-2308-10792}. This approach helps the model learn to follow instructions more reliably, improving both the accuracy of the predictions and the consistency of the generated output structure. Instruction tuning is particularly beneficial when working with small LLMs, as it allows them to leverage prior knowledge while aligning closely with task-specific expectations. We fine-tuned models on the training split of each dataset.

\paragraph{Continual-pretraining}
The most resource-intensive strategy to adapt LLMs is continual pretraining, which involves further training the model on domain-specific corpora before handling downstream tasks \citep{DBLP:journals/csur/ShiXWQWWWEW26}. Such a method presents very high data and resource requirements, making it less explored than the others described above. For this reason, we dedicate the next Section to an extensive description of our approach to it.

\subsection{Continual Pre-Training}
Continual Pre-Training (CPT) consists of further training existing LLMs on next-token prediction over a large corpus of raw textual data, aiming to better inform models on domain-specific knowledge \citep{DBLP:journals/csur/ShiXWQWWWEW26}.
In the medical context, it results in exposing models to large volumes of clinical text and biomedical literature, aiming to internalize domain-specific terminology, syntax, and reasoning patterns, resulting in more accurate and contextually appropriate outputs \citep{Luo_2022,chen2023meditron70bscalingmedicalpretraining}.
This requires significant computational resources and carefully curated data. \\
We applied this method to provide models with a comprehensive foundation of specific linguistic and terminological patterns of Italian medical text.

\paragraph{CPT data}

\begin{table}[]
\centering
\setlength{\tabcolsep}{3pt}
\begin{tabular}{l|rlc}
\toprule
\textbf{source}      & \textbf{words} &  \textbf{source} & \textbf{open}\\
\midrule
\midrule
\multicolumn{4}{c}{\textbf{\textit{scientific}}} \\
\midrule
Med Common Crawl     & 65.7M               & \textit{Med-mT5} & x \\
Drug Instructions    & 36.8M               & \textit{Med-mT5} & x \\
Medical Wikipedia    & 12.8M               & \textit{Med-mT5} & x \\
E3C corpus           & 11.6M               & \textit{Med-mT5} & x \\
WebHoseAZ            & 7.2M                & \textit{Med-mT5} & x \\
Thesis                 & 5.9M                & \textit{Med-mT5} & x \\
Medical Websites     & 3.7M                & \textit{Med-mT5} & x \\
Pubmed               & 2.4M                & \textit{Med-mT5} & x \\
Supplement descr     & 1.3M                & \textit{Med-mT5} & x \\
others               & 1.0M                & \textit{Med-mT5} & x \\
Pensiero Scientifico & 103.7M            & \textit{our} &\\
Unipd Theses         & 26.3M               & \textit{our} & x \\
Zadig                & 1.2M                & \textit{our}& \\
\midrule
\textbf{tot scientific}       & 280.1M              & & \\
\midrule
\midrule
\multicolumn{4}{c}{\textbf{\textit{clinical}}} \\
\midrule
Emergency dept.    & 125.7M              & \textit{our}& x\\
\midrule
\midrule
\textbf{tot overall}          & 405.8M              & &\\
\bottomrule
\end{tabular}
\caption{\label{tab:cpt_data} Continual Pre-Training data size by source. The \textit{scientific} dataset is composed by several sources, resulting in 278M words. The \textit{clinical} dataset is composed by documents coming from the emergency department of an Italian hospital, comprising 126M words.
Around half of the \textit{scientific} data is sourced from Med-mT5 \citep{garcia-ferrero-etal-2024-medmt5}, while the remaining is from our new data sources.
All datasets are made open-source (\textit{open}), with the exception of \textit{Pensiero Scientifico} and \textit{Zadig}, due to their restrictive licence.}
\end{table}

We used two complementary datasets (Table~\ref{tab:cpt_data}) specifically designed for domain adaptation in the Italian medical and emergency care context.
First, we collected a dataset of around 278M words (the \textbf{scientific dataset}), combining the data presented by \citet{garcia-ferrero-etal-2024-medmt5} with two new sources. 
The former is composed of multiple high-quality Italian scientific sources, including Med CommonCrawl, drug instructions, medical Wikipedia, the E3C corpus, and others.
The latter is built on two sources, namely the medicine-related thesis from the University of Padova\footnote{\url{https://thesis.unipd.it/}}, and several editions of Pensiero Scientifico and Zadig (Italian scientific publications). We obtain the thesis by scraping the web page, according to the requirements of their the CC0 licence\footnote{\url{https://thesis.unipd.it/sr/static/licenza.htm}}. We recieved Pensiero Scientifico and Zadig volumes from the editors, with a restrictive license which allows training of automatic systems but do not consent to publication of the data.\\
Then, we create a \textbf{clinical dataset} consisting of around 126M words from 1,972,254 anonymized Electronic Health Records, provided by the Fenice Network \footnote{\url{https://fenicenetwork.marionegri.it}} in collaboration with the Mario Negri Institute and San Giovanni Bosco hospital. All personal identifiers have been removed or replaced (e.g., patient names are substituted with \texttt{NOME\_PERSONA} and mobile phone numbers with \texttt{NUM\_TELEFONO}) to ensure patient privacy while preserving the clinical content and structure. This clinical dataset includes diverse document types: anamnesis, discharge letters, laboratory reports, nursing notes, radiology reports, triage assessments, clinical diaries, home-based therapy records, medical visits, and specialist consultations.
All documents are collected over a three years span at the Emergency Department of the hospital.

Both datasets undergo a different preprocessing pipelines before continual pre-training, including noise reduction, tokenization, and formatting into a unified structure suitable for language model training. The two data sources are combined during training to expose the model to both scientific literature and real-world clinical content.
The conversion from raw data to training-ready sequences follows different procedures depending on the source. For data derived from \citet{garcia-ferrero-etal-2024-medmt5}, documents are already provided in chunked textual format and can therefore be directly used after and tokenization.\\
In contrast, the Pensiero Scientifico, University of Padua theses, and Zadig datasets are originally distributed as PDF files. These documents are first converted into plain text using the Docling library \citep{Docling}, followed by cleaning procedures to remove formatting artifacts (e.g., headers, footers, page numbers, and encoding noise). The cleaned text is then segmented at the section level, with each section treated as an independent training sequence. This segmentation strategy preserves local coherence while ensuring sequences of manageable length for continual pre-training.\\

\section{Tasks and Datasets}
We address five NLP tasks defined over twelve Italian datasets and twenty sub-tasks (Table~\ref{tab:task_data}) in the medical domain, modeled according to a unified approach.
When possible, we reserve some datasets for out-of-distribution testing by not exposing models to them at training time.
Rather than designing separate structures for each task, we reformulate each problem in an instruction-following format: the model receives a textual input representing the example at hand (e.g. a clinical note, a patient report, a multiple-choice question). The model’s response is expected to be a JSON-formatted string, ensuring that the output is structured, consistent, and easily parsable by downstream systems.

\paragraph{Named Entity Recognition} Named Entity Recognition (NER) involves identifying and extracting entities in text. For this task, we based our experiments on three datasets: E3C \citep{Magnini2023}, a collection of clinical narratives; PharmaER \cite{zugarini2025pharmaer}, composed of leaflets of drugs authorized by the Italian Medicines Agency;
CardioCCC \citep{448}, a selection of cardiology clinical case reports analogous to discharge summaries.
For E3C, we defined a sub-task for each annotation type between \textit{clinical entities} and \textit{body parts}; for PharmaER we considered \textit{drugs}, \textit{diseases}, \textit{anatomical parts}, and \textit{symptoms}; for CardioCCC we considered the \textit{medications} annotations.\\
We kept the other three datasets for out-of-distribution evaluation at testing time.
To assess performances on a task identical to one present in the training data (E3C), but on different clinical notes, we consider \textit{clinical entities} in the E3C-projected version by \citet{DBLP:journals/corr/abs-2503-20568}).
To test entity types seen at training time, but with slightly different definitions due to different annotation processes, we select
\textit{diseases} enitites in health records by \citet{416} (DisteMIST).
To assess performances on entity types unseen at training type we consider
\textit{cognitive symptoms} in psychiatric records by \citet{CREMA2023104557} (PsyNIT).

\paragraph{Case Report Form Filling} Case Report Forms (CRFs) are structured documents used to systematically record patient information during clinical trials or routine care. The CRF filling task requires models to extract relevant information from unstructured clinical narratives and populate the corresponding fields in the forms. We used the \textit{diagnosis}, \textit{clinical history}, and \textit{exams} sub-tasks from the dataset proposed by \citet{ferrazzi-etal-2025-converting}, which builds on E3C by mapping clinical notes to structured CRF fields. Considering this is the only available resource for the task, we could not select any dataset for out-of-distribution evaluation.

\paragraph{Medical Question Answering} Medical Question Answering (QA) involves providing accurate answers to clinically relevant questions based on text or structured data. In this work, we focus on multiple-choice QA, where the model selects the correct answer from a set of options. We performed experiments using a dataset originating from medical exams -MedExpQA- proposed by \citet{ALONSO2024102938}. We utilized both the \textit{plain questions} sub-task and the \textit{rag} one, where each question is enriched with relevant context. \\
We kept three other datasets of medical exams for out-of-distribution evaluation at testing time:
Italian admission tests data by \citet{casola2023testing} (AT) to test on native Italian data; a translated version of MedMCQA \cite{pmlr-v174-pal22a} and MedQA \cite{medqa} proposed by \citet{DBLP:journals/corr/abs-2512-05658} to determine performances on two of the most utilized benchmarks in the field \citep{Medical-LLM-Leaderboard}.

\paragraph{Relation Extraction} Relation Extraction (RE) consists of identifying semantic relationships between medical entities in text. For this task, we leveraged the E3C dataset, focusing on the \textit{pertains-to} relations between exams and laboratory tests, and their results.
To the best of our knowledge, this is the only available resource for the task in Italian, and we could not select any dataset for out-of-distribution evaluation.

\paragraph{Argument Mining} Argument mining involves identifying and structuring reasoning elements or argumentative components in clinical text, such as claims, premises, markers, diseases, treatments, and diagnoses. We conducted experiments on the Casimedicos-Arg dataset \citep{sviridova-etal-2024-casimedicos}, which contains annotated medical texts for argument detection.
Similarly to RE and CRF filling, we could not select any dataset for out-of-distribution evaluation as Casimedicos-Arg is the only available resource for the task in the medical domain.

\paragraph{Evaluation metrics} Named Entity Recognition, Case Report Form filling, Argument Mining, and Relation Extraction are evaluated using F1 score on exact match, while multiple-choice QA is evaluated using accuracy. For each task, we report the average among datasets and subtasks. We determine an overall score by averaging individual metrics.


\begin{table}[]
\centering
\setlength{\tabcolsep}{5pt}
\begin{tabular}{llr|rrrr}
\toprule
       \multicolumn{3}{c|}{\textbf{Tasks definition}}   & \multicolumn{3}{c}{\textbf{Examples}} \\
       \midrule
\textbf{task} & \textbf{dataset} &  \textbf{n} & \textbf{train}   & \textbf{val}  & \textbf{test}\\
\midrule
  ner &cardioccc        & 1   & 250    & 100    & 150  \\
 ner &pharmaer         & 4   & 1316      & 404    & 64       \\ 
  ner  &e3c    & 2      & 451    & 67     & 628    \\
 re   &e3c    & 1     & 451    & 67     & 628    \\
 arg &casimed      & 1       & 434    & 63     & 125    \\
 crf  &e3c    & 3  & 2.170      & 404    & 2.615   \\
 qa  &medexpqa         & 2   & 434    & 63     & 125    \\
\midrule
\textbf{total} & all &  14 & 5.506 & 1.168 & 4.335 \\
\midrule
\multicolumn{6}{r}{\textit{The following are used only for o-o-d testing}} \\
\midrule
 ner & distemist     & 1 & -    & -    & 750   \\
 ner & psynit        & 1 & -    & -    & 400   \\
 ner & e3c-proj & 1 & 632  & 101  & 738   \\
 qa  & at            & 1 & 21   & -    & 500   \\
 qa  & medmcqa       & 1 & 182k & -    & 4.183 \\
 qa  & medqa         & 1 & 10k  & 1.272 & 1.273 \\
\midrule
\textbf{total} & all &  6 & - & - & 7.844 \\

\bottomrule    
\end{tabular}
\caption{Datasets selected for Italian medical NLP tasks definition: named entity recognition (\textit{ner}); relation extraction (\textit{re}); argument mining (\textit{arg}); case report forms filling (\textit{crf}); multiple-choice question answering (\textit{qa}). The number of sub-tasks defined per dataset (\textit{n}) is reported. The table reports datasets used at both training and testing time (\textit{in-distribution}), together with the ones utilized solely for testing purposes (\textit{out-of-distribution}) in the last six rows.}\label{tab:task_data}
\end{table}

\section{Experimental settings}

We implement the strategies described in Section~\ref{sec:methods} to improve the performance of small models on medical NLP tasks in Italian. 
To assess how “good” they can be, we establish competitive baselines using larger state-of-the-art LLMs. This comparison allows us to quantify the performance gap between compact and large models, and to understand under which conditions small models can offer a practical, resource-efficient alternative for real-world medical applications.

\paragraph{Baselines} We select two LLMs, Qwen3-32B and Medgemma-27B prompted via 4-shot, as baselines. Both models are chosen for their high performance on Italian general-domain tasks, following recent benchmark results \citep{DBLP:journals/corr/abs-2502-02289}. We observed that the best performances are obtained by Qwen3-32B, which we therefore considered as the primary baseline (Table~\ref{tab:small_models_results}, first two rows).

\paragraph{Models}
To select the model families for evaluation, we considered several criteria:
(i) whether the family includes models with approximately one billion parameters;
(ii) the level of adoption within the NLP research community, as we aim to provide insights that are relevant to widely used architectures; and
(iii) their performance in Italian. 
Based on these factors, we selected Llama-3.2-1B \citep{DBLP:journals/corr/abs-2407-21783}, Gemma-3-1B \citep{DBLP:journals/corr/abs-2503-19786}, and Qwen-3-1.7B \citep{DBLP:journals/corr/abs-2505-09388}. We utilized all models, obtaining six models overall.\\
Note that the way models report the number of parameters in the name differs among families: Qwen and Llama have, respectively, $72\%$ and $24\%$ more parameters than Gemma, which has an actual number of parameters exactly equal to one billion.

\paragraph{Methods to enhance performances} 
We test and evaluate the impact of several approaches to improve performance across tasks, both at inference and training time. First, we assess models in \textbf{0-shot} and \textbf{4-shot}, using structured prompts for instructed models, and much simpler ones for base models.  We then incorporate \textbf{constraint decoding}, which enforces valid output structures and reduces formatting errors during generation. 
We test the impact of \textbf{supervised fine-tuning}, which enables models to learn both the semantic aspects of the tasks and the expected output formats. Finally, we explore \textbf{continual pretraining}, further exposing models to large amounts of domain-specific medical data to enhance their understanding of specialized terminology and clinical language patterns.

\paragraph{Supervised Fine-Tuning settings}
To perform supervised fine-tuning on both the base and instruction-tuned models—as well as on their counterparts that underwent continual pretraining, 12 models in total—we employed the LoRA method \citep{DBLP:conf/iclr/HuSWALWWC22}. Fine-tuning was conducted with a learning rate of \textit{5e-4} using a cosine learning rate scheduler, on a single NVIDIA H200 GPU, with a batch size of 16. On average, training required approximately two hours per model, resulting in 24 hours overall.

\paragraph{Continual Pre-training settings}
We train using the transformers library on 3 Nvidia L40S GPUs with accelerate\footnote{\url{https://github.com/huggingface/accelerate}}. We employed $BF16$ mixed precision training to optimize memory usage while maintaining numerical stability.
In our preliminary experiments, we tested three different learning rate configurations: (1) a cosine scheduler with $30\%$ warmup and peak learning rate of $0.0002$, (2) a constant scheduler with no warmup and learning rate of $0.0002$, and (3) a constant scheduler with no warmup and lower learning rate of $5e-5$ to mitigate overfitting \cite{DBLP:journals/tmlr/IbrahimTGRABLR24}. We observed that the first configuration gives the best performance in terms of loss.\\
We utilized sequence packing techniques with Flash Attention 2 to maximize efficiency, processing sequences up to $1024$ tokens with appropriate attention masking to prevent cross-sample contamination \cite{DBLP:journals/corr/abs-2407-09105}. Gradient accumulation and synchronization were handled automatically by the Accelerate library across the 3 GPUs. Each training run required approximately 8-12 hours, depending on the model size, with larger models (Qwen-3 1.7B) taking closer to 12 hours while smaller models (Gemma-3 1B) completing in around 8 hours. 
We exposed to CPT both base and instructed-tuned models (e.g., Llama-3.2-1B and Llama-3.2-1B-Instruct), following the findings of \citet{DBLP:journals/corr/abs-2506-07597}, which highlight how reverting the order of CPT first and instruction tuning second might lead to better results. 
All models were trained on the combined scientific and clinical datasets.\\
More details about CPT choices can be found in Appendix. 

\section{Results and Discussion}

\begin{table*}[]
\centering
\small
\begin{tabular}{cl|rrrrr|>{\raggedleft\arraybackslash}p{1cm}|rl}
\toprule
\textbf{Model} & \textbf{Method} & \textbf{NER} & \textbf{CRF} & \textbf{RE} & \textbf{QA} & \textbf{ARG} & \textbf{AVG} & $\delta$ baseline & p-val \\
\midrule
 \textbf{medgemma-27b}  & \textit{+ 4-shot}    & 52.4& 62.2& 8.9& 83.2& 62.6& 53.8 \\
 \textbf{Qwen3-32B}  & \textit{+ 4-shot} & 49.7 & 63.3 & 11.5 & 82.8 & 66.3 & 54.7\\
 \midrule
\midrule
 & \textit{0-shot} & 0.0 & 0.0 & 0.0 & 0.0 & 0.0 & 0.0 & \small\textcolor{red}{-54.7} & \\
 & \textit{+ CD} & 4.8 & 0.3 & 0.0 & 16.8 & 0.0 & 4.4 & \small\textcolor{red}{-50.3} & \\
 & \textit{+ 4-shot} & 3.9 & 13.2 & 3.1 & 23.2 & 1.9 & 9.1 & \small\textcolor{red}{-45.6} & \\
 & \textit{+ CD, 4-shot} & 12.6 & 13.3 & 4.1 & 23.2 & 12.4 & 13.1 & \small\textcolor{red}{-41.6} & \\
 & \textit{+ FT} & \underline{60.1} & \underline{67.1} & 23.4 & \underline{52.4} & \underline{76.5} & 55.9 & \small\textcolor{ForestGreen}{+1.2} & * \\
\multirow{-6}{*}{\makecell[c]{\textbf{gemma-3-1b-pt}\\ \small 1.00B params}} & \textit{+ CPT, FT} & 48.0 & 42.5 & 5.6 & 19.2 & 57.5 & 34.6 & \small\textcolor{red}{-20.1} & \\

\midrule
 & \textit{0-shot} & 21.1 & 8.3 & 0.1 & 37.6 & 4.3 & 14.3 & \small\textcolor{red}{-40.4} & \\
 & \textit{+ CD} & 20.0 & 26.4 & 0.0 & 32.4 & 1.3 & 16.0 & \small\textcolor{red}{-38.7} & \\
 & \textit{+ 4-shot} & 16.9 & 25.1 & 0.0 & 32.0 & 17.8 & 18.4 & \small\textcolor{red}{-36.3} & \\
 & \textit{+ CD, 4-shot} & 16.9 & 35.8 & 0.6 & 37.2 & 30.5 & 24.2 & \small\textcolor{red}{-30.5} & \\
 & \textit{+ FT} & \textbf{61.5} & \textbf{68.5} & \underline{24.0} & 51.6 & \textbf{78.5} & \underline{56.8} & \small\textcolor{ForestGreen}{+2.1} & * \\
\multirow{-6}{*}{\makecell[c]{\textbf{gemma-3-1b-it}\\ \small 1.00B params}} & \textit{+ CPT, FT} & 53.6 & 65.8 & \textbf{47.2} & \textbf{61.8} & 68.7 & \textbf{59.4} & \small\textcolor{ForestGreen}{+4.7} & * \\
\midrule
\midrule

 & \textit{0-shot} & 0.0 & 0.0 & 0.0 & 0.0 & 0.0 & 0.0 & \small\textcolor{red}{-54.7}& \\
 & \textit{+ CD} & 6.8 & 3.3 & 0.0 & 15.6 & 1.2 & 5.4 & \small\textcolor{red}{-49.3}& \\
 & \textit{+ 4-shot} & 12.0 & 11.1 & 3.1 & 25.6 & 22.0 & 14.8 & \small\textcolor{red}{-39.9}& \\
 & \textit{+ CD, 4-shot} & 14.9 & 13.2 & 3.1 & 26.0 & 31.6 & 17.7 & \small\textcolor{red}{-37.0}& \\
 & \textit{+ FT} & 39.0 & 51.5 & \underline{30.6} & \underline{57.6} & 70.6 & 49.9 & \small\textcolor{red}{-4.8}& \\
\multirow{-6}{*}{\makecell[c]{\textbf{Llama-3.2-1B}\\ \small 1.24B params}} & \textit{+ CPT, FT} & 47.8 & 34.6 & 20.8 & 37.6 & \underline{74.8} & 43.1 & \small\textcolor{red}{-11.6} & \\
\midrule
 & \textit{0-shot} & 15.8 & 18.1 & 1.3 & 42.8 & 7.2 & 17.0 & \small\textcolor{red}{-37.7}& \\
 & \textit{+ CD} & 17.2 & 27.9 & 1.3 & 43.2 & 6.6 & 19.2 & \small\textcolor{red}{-35.5}& \\
 & \textit{+ 4-shot} & 10.3 & 23.6 & 6.3 & 36.0 & 8.7 & 17.0 & \small\textcolor{red}{-37.7}& \\
 & \textit{+ CD, 4-shot} & 11.2 & 10.5 & 6.2 & 35.6 & 23.1 & 17.3 & \small\textcolor{red}{-37.4}& \\
 & \textit{+ FT} & \underline{56.7} & \textbf{71.7} & \textbf{30.9} & \textbf{59.2} & \textbf{76.6} & \textbf{59.0} & \small\textcolor{ForestGreen}{+4.3}& **\\
\multirow{-6}{*}{\makecell[c]{\textbf{Llama-3.2-1B-Instruct}\\ \small 1.24B params}} & \textit{+ CPT, FT} & \textbf{59.7} & \underline{64.7} & 27.5 & 43.2 & 73.8 & \underline{53.8} & \small\textcolor{red}{-0.9} & \\
\midrule
\midrule

 & \textit{0-shot} & 0.0 & 0.0 & 0.0 & 0.0 & 0.0 & 0.0 & \small\textcolor{red}{-54.7}& \\
 & \textit{+ CD} & 28.6 & 7.1 & 1.0 & 7.9 & 25.8 & 14.1 & \small\textcolor{red}{-40.6}& \\
 & \textit{+ 4-shot} & 30.7 & 29.5 & 2.1 & 12.3 & 50.2 & 25.0 & \small\textcolor{red}{-29.7}& \\
 & \textit{+ CD, 4-shot} & 34.8 & 24.3 & 2.6 & 13.3 & 52.7 & 25.5 & \small\textcolor{red}{-29.2}& \\
 & \textit{+ FT} & \underline{62.2} & \textbf{73.7} & \underline{33.2} & \underline{60.8} & \underline{79.4} & \underline{61.9} & \small\textcolor{ForestGreen}{+7.2} & ***\\
\multirow{-6}{*}{\makecell[c]{\textbf{Qwen3-1.7B-Base}\\ \small 1.72B params}} & \textit{+ CPT, FT} & 59.4 & 70.3 & 18.3 & 52.8 & 74.0 & 55.0 & \small\textcolor{ForestGreen}{+0.3} & *\\
\midrule
 & \textit{0-shot} & 18.0 & 31.8 & 4.1 & 56.0 & 11.2 & 24.2 & \small\textcolor{red}{-30.5}& \\
 & \textit{+ CD} & 18.4 & 35.3 & 4.0 & 56.8 & 11.4 & 25.2 & \small\textcolor{red}{-29.5}& \\
 & \textit{+ 4-shot} & 20.6 & 37.0 & 12.3 & 44.4 & 54.5 & 33.8 & \small\textcolor{red}{-20.9}& \\
 & \textit{+ CD, 4-shot} & 20.4 & 37.1 & 12.4 & 59.2 & 54.9 & 36.8 & \small\textcolor{red}{-17.9}& \\
 & \textit{+ FT} & \textbf{62.6} & \underline{72.9} & \textbf{37.9} & \textbf{64.0} & \textbf{82.0} & \textbf{63.9} & \small\textcolor{ForestGreen}{+9.2} & ***\\
\multirow{-6}{*}{\makecell[c]{\textbf{Qwen3-1.7B}\\ \small 1.72B params}} & \textit{+ CPT, FT} & 61.5 & 67.4 & 25.3 & 57.6 & 77.6 & 57.9 & \small\textcolor{ForestGreen}{+3.2} & **\\
\bottomrule
\end{tabular}     
\caption{Impact of different methods on the five tasks for the three selected model families. For each model family, the best and second-best performances are in \textbf{bold} and \underline{underlined} respectively.
The first two rows report the performances of the baseline models Qwen3-32B and medgemma-27b-text-it.
The "$\delta$ baseline" column reports the delta from the best performing baseline (Qwen3-32B with 4-shot).
The last column (\textit{p-val)} reports the significance for the hypothesis of the adaptation method performances being higher than the baseline. One * refers to an observed significance higher than $0.7$, ** higher than $0.8$, and *** higher than $0.95$. The test is performed by pairing results at a sub-task level.}
\label{tab:small_models_results}
\end{table*}

We evaluate whether small LLMs (around $1$B parameters) can approximate the behaviour of larger models through targeted adaptation strategies. 
Table~\ref{tab:small_models_results} summarizes the performance of Gemma-3, Qwen3, and Llama-3.2 across all configurations described in Section~\ref{sec:methods}. We aggregate results by averaging the F1 scores per task (i.e., each task contributes equally to the final F1 score). In Table~\ref{tab:small_models_results_avg_methods} (Appendix) we showcase that different aggregation strategies lead to the same conclusions. We do so by averaging results per sub-task, and by assigning a weight to each sub-task according to the number of testing examples. Tables~\ref{tab:results_ner}, \ref{tab:results_other_tasks} report the results over each task.\\
To estimate the statistical significance of the observed results, we perform t-tests for paired observation, where each pair is composed by the performance of the baseline and of the adapted model, respectively. We calculate the p-value for the hypothesis of the delta in performances between the baseline and the adapted model being higher than zero, and report results in Table~\ref{tab:small_models_results}, last column.

\begin{figure}[!ht]
  \includegraphics[scale=0.28]{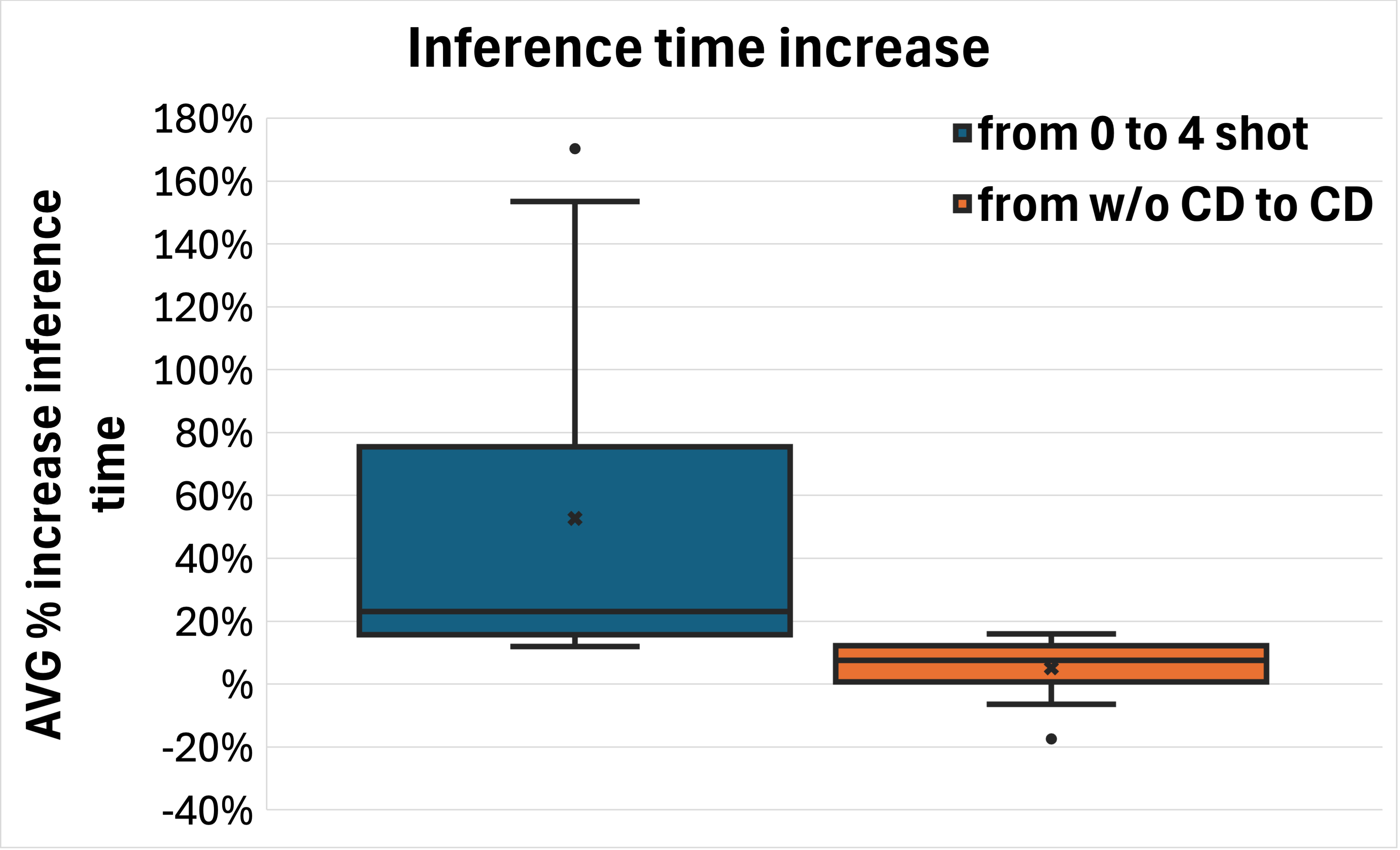}
  \caption {Impact on inference time of using \textit{4-shot} and Constraint Decoding (\textit{CD}) settings. While 4-shot significantly increases the time required to run the inference, CD does not. The average is calculated among 5 models and 14 subtasks, using the \textit{vLLM} and \textit{outlines} libraries for model serving.}
  \label{fig:inf_time}
\end{figure}

\paragraph{Inference-time methods}
We first compare inference-time methods, namely constraint decoding (CD) and few-shot prompting (4-shot), and the results are shown in Figure~\ref{fig:impact} (left).
Across all models, 4-shot prompting presents an average increase from 0-shot of $+9.4$ points, consistently yielding higher performance than constraint decoding  ($+3.8$ average increase). This difference likely arises because few-shot examples convey not only structural but also semantic cues about the task.
This gain comes at a computational cost: inference time increases by an average of 53\% when moving from 0-shot to 4-shot configurations (statistically significant), whereas applying CD has no significant impact on runtime (Figure~\ref{fig:inf_time}).
Full results in Table~\ref{tab:inf_time}).
When the context allows, combining 4-shot prompting with constraint decoding yields the most reliable results ($+13.2$ on average).\\
In general, base models show greater improvements compared to their instruction-tuned counterparts ($+9.4$ and $+3.8$ respectively), which suggests that they possess untapped potential that can be effectively steered toward task-specific behaviour through adaptation.
Interestingly, Llama-3.2-1B-Instruct shows no benefit, suggesting that it has already undergone an instruction-tuning phase closely aligned with the structure and objectives of our task definitions.


\paragraph{Training-time methods}
Among training-time strategies, supervised fine-tuning (FT) consistently produces the best results across all models and tasks (Figure~\ref{fig:impact}, right). Fine-tuning enables models to internalize both the semantic and structural aspects of the tasks, resulting in substantial performance gains.
We also evaluate continual pretraining (CPT) followed by fine-tuning. While CPT+FT achieves competitive performance and can match larger baselines, it is generally less effective than fine-tuning alone, with the notable exception of Gemma3-1b-it, where CPT provides additional benefits.

\paragraph{Task-level analysis}
A task-wise breakdown reveals consistent patterns across models. The Relation Extraction task emerges as the most challenging, indicating that smaller models still struggle with learning and reasoning over complex inter-entity relationships. In contrast, Question Answering (QA) is the easiest task, likely because models have already been exposed to large amounts of QA-style data during training. This is the only task where the baselines outperform our best adapted models.
The tasks that benefits the most from adaptation is argument mining.

\paragraph{Best models}
Our experiments show that five of the six selected small LLMs can surpass large models, demonstrating that careful adaptation can compensate for model size.
Our smallest model (Gemma-3-1b-it+CPT+FT) achieves an average score of $59.4$, $+4.7$ above the Qwen-32B, 4-shot baseline.
With $72\%$ of the parameters more than Gemma, Qwen-3-1.7B+FT stands out, reaching an overall macro score of $63.9$ and surpassing the performance of the baseline by $+9.2$ points.


\section{Out-of-distribution tasks}

\begin{table}[t]
\centering
\setlength{\tabcolsep}{5pt}
\small
\begin{tabular}{ll|lr}
\toprule
\textbf{model}        & \textbf{method} & \textbf{AVG} & \textbf{delta} \\
\midrule
Qwen3-32B (baseline)            & 4-shot          & 61.7         &                \\
\midrule
Llama-3.2-1B-Instruct & 4-shot + CD     & 24.1         &                \\
Llama-3.2-1B-Instruct & FT              & 35.9         & \textcolor{ForestGreen}{+11.9}           \\
\midrule
Qwen3-1.7B            & 4-shot + CD     & 33.2         &                \\
Qwen3-1.7B            & FT              & 39.3         & \textcolor{ForestGreen}{+6.1}            \\
\midrule
gemma-3-1b-it         & 4-shot + CD     & 27.6         &                \\
gemma-3-1b-it         & CPT + FT        & 31.4         & \textcolor{ForestGreen}{+3.9}            \\         
\bottomrule
\end{tabular}
\caption{Average performances of the adapted models and their counterparts for each of the families on Out-Of-Distribution datasets. The \textit{delta} column represents the improvement in performance due to FT/CPT.}
\label{tab:ood}
\end{table}

To assess the generalization capabilities of the trained models, we evaluate the models on six datasets not utilized at training time, highlighted in Table~\ref{tab:task_data}.
For those datasets without an official train-test split, we keep 4 examples for few shots in the baseline, and test on all the remaining ones.\\
For each model family, we select the models with the best performances in in-distribution datasets, resulting in Llama-3.2-1B-Instruct+FT, Gemma-3-1b-it+CPT+FT, and Qwen3-1.7B+FT.

To evaluate whether training improves generalization to out-of-domain datasets, we compare each adapted model (i.e., models that underwent our FT/CPT) with its corresponding pre-adaptation version. To enforce fair comparison, for each trained model, we prompt its pre-FT/CPT version using the best-performing configuration identified earlier (4-shot with constraint decoding). We evaluate the FT/CPT models in zero-shot settings to test whether the learned knowledge transfers beyond the training distribution, and compare it to the best-performing inference-time method.
The results show consistent improvements: Llama-3.2 achieves a gain of $+11.9$ points, Qwen3 improves by $+6.1$, and Gemma-3 by $+3.9$. 
Although these gains confirm that training enhances generalization, the magnitude of improvement is considerably smaller than that observed for in-distribution data.
Consequently, when compared with the Qwen3-32B 4-shot baseline, the small models still fall short in overall performance (see Table~\ref{tab:ood}, and Table~\ref{tab:results_ood_all} in Appendix for sub-task specific results).\\
In summary, while domain adaptation through fine-tuning effectively improves performance even on out-of-distribution datasets, larger models continue to generalize more robustly. These findings highlight that achieving strong generalization in specialized domains still requires task- or dataset-specific adaptation.

\section{Related Work}

\paragraph{Small Language Models} Recent work has demonstrated that SLLMs can achieve competitive performance when properly adapted to specific domains \cite{10.1145/3735633,DBLP:journals/corr/abs-2502-02737}, highlighting their efficiency as a property that makes them particularly attractive for resource-constrained environments. 
\citet{garcia-ferrero-etal-2024-medmt5, farzi-etal-2024-get} showcase how SLLMs can be effectively adapted to the medical domain.

\paragraph{Constraint decoding}
Constraint decoding consists of defining a formal grammar and enforcing LLMs to generate accordingly \citep{geng2024grammarconstraineddecodingstructurednlp}.
\citet{NEURIPS2024_2bdc2267} have presented the potential issues and how to overcome them.

\paragraph{Continual Pre-training for Domain Adaptation} Continual pre-training has emerged as an effective strategy for adapting large language models to specialized domains without losing general language understanding \cite{DBLP:journals/nn/CossuCPLTB24,DBLP:journals/tmlr/YildizRSBE25}. This approach is particularly valuable in medical domains, where specific terminology and reasoning patterns differ significantly from general text \cite{DBLP:conf/acl/GururanganMSLBD20}.

\section{Conclusion}

We found that overall, the best approach to adapt small LLMs to Italian medical tasks is to perform supervised fine-tuning, while continual pre-training rarely gives enhancements. 
If acting at inference time, combining constraint decoding with few-shot prompting offers the best results. \\
Overall, our experiments demonstrate that small LLMs can achieve competitive performance on a variety of medical NLP tasks when appropriately adapted, even exceeding larger baselines. 
Our best model, based on Qwen3-1.7B, outperforms Qwen3-32B (4-shots) by an average of $+9.2$ points. This benefit is reduced when addressing out-of-distribution data, suggesting that achieving strong performances still requires dataset-specific tuning.\\
In conclusion, we found that small, resource-efficient LLMs emerge as a viable solution for healthcare institutions with limited resources, though more research remains crucial to fully exploit their potential in out-of-distribution medical NLP tasks.

\section{Limitations}

This study focuses on Italian medical NLP datasets, which may limit the generalizability of our findings to other languages. Cross-lingual transfer and multilingual adaptation were not explored and remain avenues for future research.

Our out-of-distribution evaluation was conservative: we avoided tuning the training phase on unseen data to prevent overfitting. While this ensures a fair assessment, it may underestimate model robustness across diverse domains and institutions.

We evaluated only a limited set of adaptation strategies and did not consider reinforcement learning. Additionally, the datasets exhibit task and dataset imbalances—e.g., Question Answering is relatively easier but overrepresented, whereas Relation Extraction is underrepresented and more challenging—which may bias aggregated results. Finally, out-of-distribution evaluation does not cover all task types, limiting the completeness of our generalisation analysis across the medical NLP landscape.

\nocite{*}
\section{Bibliographical References}\label{sec:reference}

\bibliographystyle{lrec-coling2024-natbib}
\bibliography{lrec-coling2024-example}

@Inbook{Magnini2023,
author = {Magnini, Bernardo and Altuna, Bego{\~{n}}a and Lavelli, Alberto and Minard, Anne-Lyse and Speranza, Manuela and Zanoli, Roberto},
editor="Rehm, Georg",
title="European Clinical Case Corpus",
bookTitle="European Language Grid: A Language Technology Platform for Multilingual Europe",
year="2023",
publisher="Springer International Publishing",
address="Cham",
pages="283--288",
isbn="978-3-031-17258-8",
url="https://doi.org/10.1007/978-3-031-17258-8_17"
}

@article{ALONSO2024102938,
title = {{MedExpQA}: Multilingual benchmarking of Large Language Models for Medical Question Answering},
journal = {Artificial Intelligence in Medicine},
volume = {155},
pages = {102938},
year = {2024},
issn = {0933-3657},
doi = {https://doi.org/10.1016/j.artmed.2024.102938},
url = {https://www.sciencedirect.com/science/article/pii/S0933365724001805},
author = {Iñigo Alonso and Maite Oronoz and Rodrigo Agerri}
}

@inproceedings{sviridova-etal-2024-casimedicos,
    title = {{CasiMedicos-Arg: A Medical Question Answering Dataset Annotated with Explanatory Argumentative Structures}},
    author = "Sviridova, Ekaterina  and
      Yeginbergen, Anar  and
      Estarrona, Ainara  and
      Cabrio, Elena  and
      Villata, Serena  and
      Agerri, Rodrigo",
    booktitle = "Proceedings of the 2024 Conference on Empirical Methods in Natural Language Processing",
    year = "2024",
    url = "https://aclanthology.org/2024.emnlp-main.1026",
    pages = "18463--18475"
}

@conference {448,
	title = {Overview of {MultiCardioNER} task at {BioASQ} 2024 on Medical Specialty and Language Adaptation of Clinical NER Systems for {Spanish}, {English} and {Italian}},
	booktitle = {Conference and Labs of the Evaluation Forum},
	year = {2024},
	url = {https://ceur-ws.org/Vol-3740/paper-02.pdf},
	author = {Salvador Lima-L{\'o}pez and Eul{\`a}lia Farr{\'e}-Maduell and Jan Rodr{\'\i}guez-Miret and Miguel Rodr{\'\i}guez-Ortega and Livia Lilli and Jacopo Lenkowicz and Giovanna Ceroni and Jonathan Kossoff and Anoop Shah and Anastasios Nentidis and Anastasia Krithara and Georgios Katsimpras and Georgios Paliouras and Martin Krallinger}
}

@article{DBLP:journals/corr/abs-2502-02289,
  author       = {Bernardo Magnini and
                  Roberto Zanoli and
                  Michele Resta and
                  Martin Cimmino and
                  Paolo Albano and
                  Marco Madeddu and
                  Viviana Patti},
  title        = {{Evalita-LLM}: Benchmarking Large Language Models on {Italian}},
  journal      = {CoRR},
  volume       = {abs/2502.02289},
  year         = {2025},
  url          = {https://doi.org/10.48550/arXiv.2502.02289},
  doi          = {10.48550/ARXIV.2502.02289},
  eprinttype    = {arXiv},
  eprint       = {2502.02289},
  bibsource    = {dblp computer science bibliography, https://dblp.org}
}

@article{DBLP:journals/corr/abs-2503-19786,
  author       = {Gemma Team},
  title        = {Gemma 3 Technical Report},
  journal      = {CoRR},
  volume       = {abs/2503.19786},
  year         = {2025},
  url          = {https://doi.org/10.48550/arXiv.2503.19786},
  doi          = {10.48550/ARXIV.2503.19786},
  eprinttype    = {arXiv},
  eprint       = {2503.19786},
  bibsource    = {dblp computer science bibliography, https://dblp.org}
}

@article{DBLP:journals/corr/abs-2505-09388,
  author       = {An Yang and
                  Anfeng Li and
                  Baosong Yang and
                  Beichen Zhang and
                  Binyuan Hui and
                  Bo Zheng and
                  Bowen Yu and
                  Chang Gao and
                  Chengen Huang and
                  Chenxu Lv and
                  Chujie Zheng and
                  Dayiheng Liu and
                  Fan Zhou and
                  Fei Huang and
                  Feng Hu and
                  Hao Ge and
                  Haoran Wei and
                  Huan Lin and
                  Jialong Tang and
                  Jian Yang and
                  Jianhong Tu and
                  Jianwei Zhang and
                  Jian Yang and
                  Jiaxi Yang and
                  Jingren Zhou and
                  Junyang Lin and
                  Kai Dang and
                  Keqin Bao and
                  Kexin Yang and
                  Le Yu and
                  Lianghao Deng and
                  Mei Li and
                  Mingfeng Xue and
                  Mingze Li and
                  Pei Zhang and
                  Peng Wang and
                  Qin Zhu and
                  Rui Men and
                  Ruize Gao and
                  Shixuan Liu and
                  Shuang Luo and
                  Tianhao Li and
                  Tianyi Tang and
                  Wenbiao Yin and
                  Xingzhang Ren and
                  Xinyu Wang and
                  Xinyu Zhang and
                  Xuancheng Ren and
                  Yang Fan and
                  Yang Su and
                  Yichang Zhang and
                  Yinger Zhang and
                  Yu Wan and
                  Yuqiong Liu and
                  Zekun Wang and
                  Zeyu Cui and
                  Zhenru Zhang and
                  Zhipeng Zhou and
                  Zihan Qiu},
  title        = {Qwen3 Technical Report},
  journal      = {CoRR},
  volume       = {abs/2505.09388},
  year         = {2025},
  url          = {https://doi.org/10.48550/arXiv.2505.09388},
  doi          = {10.48550/ARXIV.2505.09388},
  eprinttype    = {arXiv},
  eprint       = {2505.09388},
  bibsource    = {dblp computer science bibliography, https://dblp.org}
}

@article{DBLP:journals/corr/abs-2407-21783,
  author       = {Abhimanyu Dubey and
                  Abhinav Jauhri and
                  Abhinav Pandey and
                  Abhishek Kadian and
                  Ahmad Al{-}Dahle and
                  Aiesha Letman and
                  Akhil Mathur and
                  Alan Schelten and
                  Amy Yang and
                  Angela Fan and
                  Anirudh Goyal and
                  Anthony Hartshorn and
                  Aobo Yang and
                  Archi Mitra and
                  Archie Sravankumar and
                  Artem Korenev and
                  Arthur Hinsvark and
                  Arun Rao and
                  Aston Zhang and
                  Aur{\'{e}}lien Rodriguez and
                  Austen Gregerson and
                  Ava Spataru and
                  Baptiste Rozi{\`{e}}re and
                  Bethany Biron and
                  Binh Tang and
                  Bobbie Chern and
                  Charlotte Caucheteux and
                  Chaya Nayak and
                  Chloe Bi and
                  Chris Marra and
                  Chris McConnell and
                  Christian Keller and
                  Christophe Touret and
                  Chunyang Wu and
                  Corinne Wong and
                  Cristian Canton Ferrer and
                  Cyrus Nikolaidis and
                  Damien Allonsius and
                  Daniel Song and
                  Danielle Pintz and
                  Danny Livshits and
                  David Esiobu and
                  Dhruv Choudhary and
                  Dhruv Mahajan and
                  Diego Garcia{-}Olano and
                  Diego Perino and
                  Dieuwke Hupkes and
                  Egor Lakomkin and
                  Ehab AlBadawy and
                  Elina Lobanova and
                  Emily Dinan and
                  Eric Michael Smith and
                  Filip Radenovic and
                  Frank Zhang and
                  Gabriel Synnaeve and
                  Gabrielle Lee and
                  Georgia Lewis Anderson and
                  Graeme Nail and
                  Gr{\'{e}}goire Mialon and
                  Guan Pang and
                  Guillem Cucurell and
                  Hailey Nguyen and
                  Hannah Korevaar and
                  Hu Xu and
                  Hugo Touvron and
                  Iliyan Zarov and
                  Imanol Arrieta Ibarra and
                  Isabel M. Kloumann and
                  Ishan Misra and
                  Ivan Evtimov and
                  Jade Copet and
                  Jaewon Lee and
                  Jan Geffert and
                  Jana Vranes and
                  Jason Park and
                  Jay Mahadeokar and
                  Jeet Shah and
                  Jelmer van der Linde and
                  Jennifer Billock and
                  Jenny Hong and
                  Jenya Lee and
                  Jeremy Fu and
                  Jianfeng Chi and
                  Jianyu Huang and
                  Jiawen Liu and
                  Jie Wang and
                  Jiecao Yu and
                  Joanna Bitton and
                  Joe Spisak and
                  Jongsoo Park and
                  Joseph Rocca and
                  Joshua Johnstun and
                  Joshua Saxe and
                  Junteng Jia and
                  Kalyan Vasuden Alwala and
                  Kartikeya Upasani and
                  Kate Plawiak and
                  Ke Li and
                  Kenneth Heafield and
                  Kevin Stone and
                  et al.},
  title        = {The {Llama} 3 Herd of Models},
  journal      = {CoRR},
  volume       = {abs/2407.21783},
  year         = {2024},
  url          = {https://doi.org/10.48550/arXiv.2407.21783},
  doi          = {10.48550/ARXIV.2407.21783},
  eprinttype    = {arXiv},
  eprint       = {2407.21783},
  bibsource    = {dblp computer science bibliography, https://dblp.org}
}

@article{DBLP:journals/csur/ShiXWQWWWEW26,
  author       = {Haizhou Shi and
                  Zihao Xu and
                  Hengyi Wang and
                  Weiyi Qin and
                  Wenyuan Wang and
                  Yibin Wang and
                  Zifeng Wang and
                  Sayna Ebrahimi and
                  Hao Wang},
  title        = {Continual Learning of Large Language Models: {A} Comprehensive Survey},
  journal      = {{ACM} Comput. Surv.},
  volume       = {58},
  number       = {5},
  year         = {2026},
  url          = {https://doi.org/10.1145/3735633}
}

@inproceedings{DBLP:conf/nips/BrownMRSKDNSSAA20,
  author       = {Tom B. Brown and
                  Benjamin Mann and
                  Nick Ryder and
                  Melanie Subbiah and
                  Jared Kaplan and
                  Prafulla Dhariwal and
                  Arvind Neelakantan and
                  Pranav Shyam and
                  Girish Sastry and
                  Amanda Askell and
                  Sandhini Agarwal and
                  Ariel Herbert{-}Voss and
                  Gretchen Krueger and
                  Tom Henighan and
                  Rewon Child and
                  Aditya Ramesh and
                  Daniel M. Ziegler and
                  Jeffrey Wu and
                  Clemens Winter and
                  Christopher Hesse and
                  Mark Chen and
                  Eric Sigler and
                  Mateusz Litwin and
                  Scott Gray and
                  Benjamin Chess and
                  Jack Clark and
                  Christopher Berner and
                  Sam McCandlish and
                  Alec Radford and
                  Ilya Sutskever and
                  Dario Amodei},
  editor       = {Hugo Larochelle and
                  Marc'Aurelio Ranzato and
                  Raia Hadsell and
                  Maria{-}Florina Balcan and
                  Hsuan{-}Tien Lin},
  title        = {Language Models are Few-Shot Learners},
  booktitle    = {Advances in Neural Information Processing Systems 33: Annual Conference
                  on Neural Information Processing Systems 2020, NeurIPS 2020, December
                  6-12, 2020, virtual},
  year         = {2020},
  url          = {https://proceedings.neurips.cc/paper/2020/hash/1457c0d6bfcb4967418bfb8ac142f64a-Abstract.html}
}

@article{DBLP:journals/corr/abs-2308-10792,
author = {Zhang, Shengyu and Dong, Linfeng and Li, Xiaoya and Zhang, Sen and Sun, Xiaofei and Wang, Shuhe and Li, Jiwei and Hu, Runyi and Zhang, Tianwei and Wang, Guoyin and Wu, Fei},
title = {Instruction Tuning for Large Language Models: A Survey},
year = {2026},
issue_date = {May 2026},
publisher = {Association for Computing Machinery},
address = {New York, NY, USA},
volume = {58},
number = {7},
issn = {0360-0300},
url = {https://doi.org/10.1145/3777411},
doi = {10.1145/3777411},
journal = {ACM Comput. Surv.},
month = jan,
articleno = {169},
numpages = {36}
}

@inproceedings{DBLP:conf/iclr/HuSWALWWC22,
  author       = {Edward J. Hu and
                  Yelong Shen and
                  Phillip Wallis and
                  Zeyuan Allen{-}Zhu and
                  Yuanzhi Li and
                  Shean Wang and
                  Lu Wang and
                  Weizhu Chen},
  title        = {{LoRA}: Low-Rank Adaptation of Large Language Models},
  booktitle    = {The Tenth International Conference on Learning Representations, {ICLR}
                  2022, Virtual Event, April 25-29, 2022},
  publisher    = {OpenReview.net},
  year         = {2022},
  url          = {https://openreview.net/forum?id=nZeVKeeFYf9},
  bibsource    = {dblp computer science bibliography, https://dblp.org}
}

@ARTICLE{10651607,
  author={Crema, Claudio and Verde, Federico and Tiraboschi, Pietro and Marra, Camillo and Arighi, Andrea and Fostinelli, Silvia and Giuffré, Guido Maria and Maschio, Vera Pacoova Dal and L'Abbate, Federica and Solca, Federica and Poletti, Barbara and Silani, Vincenzo and Rotondo, Emanuela and Borracci, Vittoria and Vimercati, Roberto and Crepaldi, Valeria and Inguscio, Emanuela and Filippi, Massimo and Caso, Francesca and Rosati, Alessandra Maria and Quaranta, Davide and Binetti, Giuliano and Pagnoni, Ilaria and Morreale, Manuela and Burgio, Francesca and Stanzani-Maserati, Michelangelo and Capellari, Sabina and Pardini, Matteo and Girtler, Nicola and Piras, Federica and Piras, Fabrizio and Lalli, Stefania and Perdixi, Elena and Lombardi, Gemma and Tella, Sonia Di and Costa, Alfredo and Capelli, Marco and Fundarò, Cira and Manera, Marina and Muscio, Cristina and Pellencin, Elisa and Lodi, Raffaele and Tagliavini, Fabrizio and Redolfi, Alberto},
  journal={IEEE Journal of Biomedical and Health Informatics}, 
  title={Medical Information Extraction With {NLP}-Powered {QABots}: A Real-World Scenario}, 
  year={2024},
  volume={28},
  number={11},
  pages={6906-6917},
  doi={10.1109/JBHI.2024.3450118}
}

@inproceedings{garcia-ferrero-etal-2024-medmt5,
    title = "{M}ed{MT}5: An Open-Source Multilingual Text-to-Text {LLM} for the Medical Domain",
    author = "Garc{\'i}a-Ferrero, Iker  and
      Agerri, Rodrigo  and
      Atutxa Salazar, Aitziber  and
      Cabrio, Elena  and
      de la Iglesia, Iker  and
      Lavelli, Alberto  and
      Magnini, Bernardo  and
      Molinet, Benjamin  and
      Ramirez-Romero, Johana  and
      Rigau, German  and
      Villa-Gonzalez, Jose Maria  and
      Villata, Serena  and
      Zaninello, Andrea",
    editor = "Calzolari, Nicoletta  and
      Kan, Min-Yen  and
      Hoste, Veronique  and
      Lenci, Alessandro  and
      Sakti, Sakriani  and
      Xue, Nianwen",
    booktitle = "Proceedings of the 2024 Joint International Conference on Computational Linguistics, Language Resources and Evaluation (LREC-COLING 2024)",
    month = may,
    year = "2024",
    address = "Torino, Italia",
    publisher = "ELRA and ICCL",
    url = "https://aclanthology.org/2024.lrec-main.974/",
}

@inproceedings{ferrazzi-etal-2025-converting,
    title = "Converting Annotated Clinical Cases into Structured Case Report Forms",
    author = "Ferrazzi, Pietro  and
      Lavelli, Alberto  and
      Magnini, Bernardo",
    editor = "Demner-Fushman, Dina  and
      Ananiadou, Sophia  and
      Miwa, Makoto  and
      Tsujii, Junichi",
    booktitle = "Proceedings of the 24th Workshop on Biomedical Language Processing",
    month = aug,
    year = "2025",
    address = "Vienna, Austria",
    publisher = "Association for Computational Linguistics",
    url = "https://aclanthology.org/2025.bionlp-1.26/",
    doi = "10.18653/v1/2025.bionlp-1.26",
}

@inproceedings{DBLP:journals/corr/abs-2506-07597,
    title = "Instructing Large Language Models for Low-Resource Languages: A Systematic Study for {B}asque",
    author = "Sainz, Oscar  and
      Perez, Naiara  and
      Etxaniz, Julen  and
      Fernandez de Landa, Joseba  and
      Aldabe, Itziar  and
      Garc{\'i}a-Ferrero, Iker  and
      Zabala, Aimar  and
      Azurmendi, Ekhi  and
      Rigau, German  and
      Agirre, Eneko  and
      Artetxe, Mikel  and
      Soroa, Aitor",
    editor = "Christodoulopoulos, Christos  and
      Chakraborty, Tanmoy  and
      Rose, Carolyn  and
      Peng, Violet",
    booktitle = "Proceedings of the 2025 Conference on Empirical Methods in Natural Language Processing",
    month = nov,
    year = "2025",
    address = "Suzhou, China",
    publisher = "Association for Computational Linguistics",
    url = "https://aclanthology.org/2025.emnlp-main.1484/",
    doi = "10.18653/v1/2025.emnlp-main.1484",
    pages = "29136--29160",
}

@inproceedings{casola2023testing,
  title={Testing {ChatGPT} for Stability and Reasoning: A Case Study Using {I}talian Medical Specialty Tests.},
  author={Casola, Silvia and Labruna, Tiziano and Lavelli, Alberto and Magnini, Bernardo and others},
  booktitle={Proceedings of the Nineth Italian Conference on Computational Linguistics},
  year={2023}
}

@article{DBLP:journals/corr/abs-2503-20568,
  author       = {Soumitra Ghosh and
                  Bego{\~{n}}a Altuna and
                  Saeed Farzi and
                  Pietro Ferrazzi and
                  Alberto Lavelli and
                  Giulia Mezzanotte and
                  Manuela Speranza and
                  Bernardo Magnini},
  title        = {Low-resource Information Extraction with the {European Clinical Case
                  Corpus}},
  journal      = {CoRR},
  volume       = {abs/2503.20568},
  year         = {2025},
  url          = {https://doi.org/10.48550/arXiv.2503.20568},
  doi          = {10.48550/ARXIV.2503.20568},
  eprinttype    = {arXiv},
  eprint       = {2503.20568},
  bibsource    = {dblp computer science bibliography, https://dblp.org}
}

@inproceedings{farzi-etal-2024-get,
    title = "Get the Best out of 1{B} {LLM}s: Insights from Information Extraction on Clinical Documents",
    author = "Farzi, Saeed  and
      Ghosh, Soumitra  and
      Lavelli, Alberto  and
      Magnini, Bernardo",
    editor = "Demner-Fushman, Dina  and
      Ananiadou, Sophia  and
      Miwa, Makoto  and
      Roberts, Kirk  and
      Tsujii, Junichi",
    booktitle = "Proceedings of the 23rd Workshop on Biomedical Natural Language Processing",
    month = aug,
    year = "2024",
    address = "Bangkok, Thailand",
    publisher = "Association for Computational Linguistics",
    url = "https://aclanthology.org/2024.bionlp-1.21/",
    doi = "10.18653/v1/2024.bionlp-1.21",
}

@InProceedings{pmlr-v174-pal22a,
  title = 	 {{MedMCQA}: A Large-scale Multi-Subject Multi-Choice Dataset for Medical domain Question Answering},
  author =       {Pal, Ankit and Umapathi, Logesh Kumar and Sankarasubbu, Malaikannan},
  booktitle = 	 {Proceedings of the Conference on Health, Inference, and Learning},
  pages = 	 {248--260},
  year = 	 {2022},
  editor = 	 {Flores, Gerardo and Chen, George H and Pollard, Tom and Ho, Joyce C and Naumann, Tristan},
  volume = 	 {174},
  series = 	 {Proceedings of Machine Learning Research},
  month = 	 {07--08 Apr},
  publisher =    {PMLR},
  url = 	 {https://proceedings.mlr.press/v174/pal22a.html},
}

@article{medqa,
author = {Jin, Di and Pan, Eileen and Oufattole, Nassim and Weng, Wei-Hung and Fang, Hanyi and Szolovits, Peter},
year = {2021},
month = {07},
pages = {6421},
title = {What Disease Does This Patient Have? A Large-Scale Open Domain Question Answering Dataset from Medical Exams},
volume = {11},
journal = {Applied Sciences},
doi = {10.3390/app11146421}
}

@article{10.1145/3735633,
author = {Shi, Haizhou and Xu, Zihao and Wang, Hengyi and Qin, Weiyi and Wang, Wenyuan and Wang, Yibin and Wang, Zifeng and Ebrahimi, Sayna and Wang, Hao},
title = {Continual Learning of Large Language Models: A Comprehensive Survey},
year = {2025},
issue_date = {April 2026},
publisher = {Association for Computing Machinery},
address = {New York, NY, USA},
volume = {58},
number = {5},
issn = {0360-0300},
url = {https://doi.org/10.1145/3735633},
doi = {10.1145/3735633},
journal = {ACM Comput. Surv.},
month = nov,
articleno = {120},
numpages = {42}
}

@inproceedings{
DBLP:journals/corr/abs-2502-02737,
title={Smol{LM}2: When Smol Goes Big {\textemdash} Data-Centric Training of a Fully Open Small Language Model},
author={Loubna Ben allal and Anton Lozhkov and Elie Bakouch and Gabriel Martin Blazquez and Guilherme Penedo and Lewis Tunstall and Andr{\'e}s Marafioti and Agust{\'\i}n Piqueres Lajar{\'\i}n and Hynek Kydl{\'\i}{\v{c}}ek and Vaibhav Srivastav and Joshua Lochner and Caleb Fahlgren and Xuan Son NGUYEN and Ben Burtenshaw and Cl{\'e}mentine Fourrier and Haojun Zhao and Hugo Larcher and Mathieu Morlon and Cyril Zakka and Colin Raffel and Leandro Von Werra and Thomas Wolf},
booktitle={Second Conference on Language Modeling},
year={2025},
url={https://openreview.net/forum?id=3JiCl2A14H}
}

@article{DBLP:journals/nn/CossuCPLTB24,
  author       = {Andrea Cossu and
                  Antonio Carta and
                  Lucia C. Passaro and
                  Vincenzo Lomonaco and
                  Tinne Tuytelaars and
                  Davide Bacciu},
  title        = {Continual pre-training mitigates forgetting in language and vision},
  journal      = {Neural Networks},
  volume       = {179},
  pages        = {106492},
  year         = {2024},
  url          = {https://doi.org/10.1016/j.neunet.2024.106492},
  doi          = {10.1016/J.NEUNET.2024.106492},
  bibsource    = {dblp computer science bibliography, https://dblp.org}
}

@article{DBLP:journals/tmlr/YildizRSBE25,
  author       = {{\c{C}}agatay Yildiz and
                  Nishaanth Kanna Ravichandran and
                  Nitin Sharma and
                  Matthias Bethge and
                  Beyza Ermis},
  title        = {Investigating Continual Pretraining in Large Language Models: Insights
                  and Implications},
  journal      = {Trans. Mach. Learn. Res.},
  volume       = {2025},
  year         = {2025},
  url          = {https://openreview.net/forum?id=aKjJoEVKgO},
  bibsource    = {dblp computer science bibliography, https://dblp.org}
}

@inproceedings{DBLP:conf/acl/GururanganMSLBD20,
  author       = {Suchin Gururangan and
                  Ana Marasovic and
                  Swabha Swayamdipta and
                  Kyle Lo and
                  Iz Beltagy and
                  Doug Downey and
                  Noah A. Smith},
  editor       = {Dan Jurafsky and
                  Joyce Chai and
                  Natalie Schluter and
                  Joel R. Tetreault},
  title        = {Don't Stop Pretraining: Adapt Language Models to Domains and Tasks},
  booktitle    = {Proceedings of the 58th Annual Meeting of the Association for Computational
                  Linguistics, {ACL} 2020, Online, July 5-10, 2020},
  pages        = {8342--8360},
  publisher    = {Association for Computational Linguistics},
  year         = {2020},
  url          = {https://doi.org/10.18653/v1/2020.acl-main.740},
  doi          = {10.18653/V1/2020.ACL-MAIN.740},
  bibsource    = {dblp computer science bibliography, https://dblp.org}
}

@article{DBLP:journals/corr/abs-2407-09105,
  author       = {Achintya Kundu and
                  Rhui Dih Lee and
                  Laura Wynter and
                  Raghu Kiran Ganti and
                  Mayank Mishra},
  title        = {Enhancing Training Efficiency Using Packing with Flash Attention},
  journal      = {CoRR},
  volume       = {abs/2407.09105},
  year         = {2024},
  url          = {https://doi.org/10.48550/arXiv.2407.09105},
  doi          = {10.48550/ARXIV.2407.09105},
  eprinttype    = {arXiv},
  eprint       = {2407.09105},
  bibsource    = {dblp computer science bibliography, https://dblp.org}
}

@article{DBLP:journals/tmlr/IbrahimTGRABLR24,
  author       = {Adam Ibrahim and
                  Benjamin Th{\'{e}}rien and
                  Kshitij Gupta and
                  Mats L. Richter and
                  Quentin Gregory Anthony and
                  Eugene Belilovsky and
                  Timoth{\'{e}}e Lesort and
                  Irina Rish},
  title        = {Simple and Scalable Strategies to Continually Pre-train Large Language
                  Models},
  journal      = {Trans. Mach. Learn. Res.},
  volume       = {2024},
  year         = {2024},
  url          = {https://openreview.net/forum?id=DimPeeCxKO},
  bibsource    = {dblp computer science bibliography, https://dblp.org}
}

@inproceedings{zugarini2025pharmaer,
  title={{PharmaER.IT}: an {Italian} Dataset for Entity Recognition in the Pharmaceutical Domain},
  author={Zugarini, Andrea and Rigutini, Leonardo},
  year={2025},
  booktitle={Proceedings of the Eleventh Italian Conference on Computational Linguistics}, 
}

@article{CREMA2023104557,
title = {Advancing Italian biomedical information extraction with transformers-based models: Methodological insights and multicenter practical application},
journal = {Journal of Biomedical Informatics},
volume = {148},
pages = {104557},
year = {2023},
issn = {1532-0464},
doi = {https://doi.org/10.1016/j.jbi.2023.104557},
url = {https://www.sciencedirect.com/science/article/pii/S1532046423002782},
author = {Claudio Crema and Tommaso Mario Buonocore and Silvia Fostinelli and Enea Parimbelli and Federico Verde and Cira Fundarò and Marina Manera and Matteo Cotta Ramusino and Marco Capelli and Alfredo Costa and Giuliano Binetti and Riccardo Bellazzi and Alberto Redolfi}
}

@conference {416,
	title = {Overview of {DisTEMIST} at {BioASQ}: Automatic detection and normalization of diseases from clinical texts: results, methods, evaluation and multilingual resources},
	booktitle = {Working Notes of Conference and Labs of the Evaluation (CLEF) Forum. CEUR Workshop Proceedings},
	year = {2022},
	url = {https://ceur-ws.org/Vol-3180/paper-11.pdf},
	author = {Miranda-Escalada, Antonio and Gasco, Luis and Lima-L{\'o}pez, Salvador and Farr{\'e}-Maduell, Eul{\`a}lia and Estrada, Darryl and Nentidis, Anastasios and Krithara, Anastasia and Katsimpras, Georgios and Paliouras, Georgios and Krallinger, Martin}
}

@inproceedings{10.1145/3711896.3736563,
author = {Wang, Fali and Lin, Minhua and Ma, Yao and Liu, Hui and He, Qi and Tang, Xianfeng and Tang, Jiliang and Pei, Jian and Wang, Suhang},
title = {A Survey on Small Language Models in the Era of Large Language Models: Architecture, Capabilities, and Trustworthiness},
year = {2025},
isbn = {9798400714542},
publisher = {Association for Computing Machinery},
address = {New York, NY, USA},
url = {https://doi.org/10.1145/3711896.3736563},
doi = {10.1145/3711896.3736563},
booktitle = {Proceedings of the 31st ACM SIGKDD Conference on Knowledge Discovery and Data Mining V.2},
pages = {6173–6183},
numpages = {11},
location = {Toronto ON, Canada},
series = {KDD '25}
}

@inproceedings{geng2024grammarconstraineddecodingstructurednlp,
  author       = {Saibo Geng and
                  Martin Josifoski and
                  Maxime Peyrard and
                  Robert West},
  editor       = {Houda Bouamor and
                  Juan Pino and
                  Kalika Bali},
  title        = {Grammar-Constrained Decoding for Structured {NLP} Tasks without Finetuning},
  booktitle    = {Proceedings of the 2023 Conference on Empirical Methods in Natural
                  Language Processing, {EMNLP} 2023, Singapore, December 6-10, 2023},
  pages        = {10932--10952},
  publisher    = {Association for Computational Linguistics},
  year         = {2023},
  url          = {https://doi.org/10.18653/v1/2023.emnlp-main.674},
  doi          = {10.18653/V1/2023.EMNLP-MAIN.674},
  bibsource    = {dblp computer science bibliography, https://dblp.org}
}

@inproceedings{NEURIPS2024_2bdc2267,
 author = {Park, Kanghee and Wang, Jiayu and Berg-Kirkpatrick, Taylor and Polikarpova, Nadia and D\textquotesingle Antoni, Loris},
 booktitle = {Advances in Neural Information Processing Systems},
 editor = {A. Globerson and L. Mackey and D. Belgrave and A. Fan and U. Paquet and J. Tomczak and C. Zhang},
 pages = {24547--24568},
 publisher = {Curran Associates, Inc.},
 title = {Grammar-Aligned Decoding},
 url = {https://proceedings.neurips.cc/paper_files/paper/2024/file/2bdc2267c3d7d01523e2e17ac0a754f3-Paper-Conference.pdf},
 volume = {37},
 year = {2024}
}

@misc{kundu2024enhancingtrainingefficiencyusing,
      title={Enhancing Training Efficiency Using Packing with Flash Attention}, 
      author={Achintya Kundu and Rhui Dih Lee and Laura Wynter and Raghu Kiran Ganti and Mayank Mishra},
      year={2024},
      eprint={2407.09105},
      archivePrefix={arXiv},
      primaryClass={cs.LG},
      url={https://arxiv.org/abs/2407.09105}, 
}

@inproceedings{DBLP:conf/nips/DaoFERR22,
  author       = {Tri Dao and
                  Daniel Y. Fu and
                  Stefano Ermon and
                  Atri Rudra and
                  Christopher R{\'{e}}},
  editor       = {Sanmi Koyejo and
                  S. Mohamed and
                  A. Agarwal and
                  Danielle Belgrave and
                  K. Cho and
                  A. Oh},
  title        = {FlashAttention: Fast and Memory-Efficient Exact Attention with IO-Awareness},
  booktitle    = {Advances in Neural Information Processing Systems 35: Annual Conference
                  on Neural Information Processing Systems 2022, NeurIPS 2022, New Orleans,
                  LA, USA, November 28 - December 9, 2022},
  year         = {2022},
  bibsource    = {dblp computer science bibliography, https://dblp.org}
}

@inproceedings{DBLP:conf/iclr/Dao24,
  author       = {Tri Dao},
  title        = {FlashAttention-2: Faster Attention with Better Parallelism and Work
                  Partitioning},
  booktitle    = {The Twelfth International Conference on Learning Representations,
                  {ICLR} 2024, Vienna, Austria, May 7-11, 2024},
  publisher    = {OpenReview.net},
  year         = {2024},
  url          = {https://openreview.net/forum?id=mZn2Xyh9Ec},
  bibsource    = {dblp computer science bibliography, https://dblp.org}
}

@misc{ferrazzi2025groundedmultilingualmedicalreasoning,
      title={Grounded Multilingual Medical Reasoning for Question Answering with Large Language Models}, 
      author={Pietro Ferrazzi and Aitor Soroa and Rodrigo Agerri},
      year={2025},
      eprint={2512.05658},
      archivePrefix={arXiv},
      primaryClass={cs.CL},
      url={https://arxiv.org/abs/2512.05658}, 
}

@article{DBLP:journals/corr/abs-2512-05658,
  author       = {Pietro Ferrazzi and
                  Aitor Soroa and
                  Rodrigo Agerri},
  title        = {Grounded Multilingual Medical Reasoning for Question Answering with
                  Large Language Models},
  journal      = {CoRR},
  volume       = {abs/2512.05658},
  year         = {2025},
  url          = {https://doi.org/10.48550/arXiv.2512.05658},
  doi          = {10.48550/ARXIV.2512.05658},
  eprinttype    = {arXiv},
  eprint       = {2512.05658},
  bibsource    = {dblp computer science bibliography, https://dblp.org}
}

@techreport{Docling,
  author = {Deep Search Team},
  month = {8},
  title = {Docling Technical Report},
  url = {https://arxiv.org/abs/2408.09869},
  eprint = {2408.09869},
  doi = {10.48550/arXiv.2408.09869},
  version = {1.0.0},
  year = {2024}
}

@misc{Medical-LLM-Leaderboard,
author = {Ankit Pal and Pasquale Minervini and Andreas Geert Motzfeldt and Aryo Pradipta Gema and Beatrice Alex},
title = {openlifescienceai/open-medical-llm-leaderboard},
year = {2024},
publisher = {Hugging Face},
howpublished = "{https://huggingface.co/spaces/openlifescienceaiopen-medical-llm-leaderboard}"
}

@article{Luo_2022,
   title={BioGPT: generative pre-trained transformer for biomedical text generation and mining},
   volume={23},
   ISSN={1477-4054},
   url={http://dx.doi.org/10.1093/bib/bbac409},
   DOI={10.1093/bib/bbac409},
   number={6},
   journal={Briefings in Bioinformatics},
   publisher={Oxford University Press (OUP)},
   author={Luo, Renqian and Sun, Liai and Xia, Yingce and Qin, Tao and Zhang, Sheng and Poon, Hoifung and Liu, Tie-Yan},
   year={2022},
   month=sep }

@misc{chen2023meditron70bscalingmedicalpretraining,
      title={MEDITRON-70B: Scaling Medical Pretraining for Large Language Models}, 
      author={Zeming Chen and Alejandro Hernández Cano and Angelika Romanou and Antoine Bonnet and Kyle Matoba and Francesco Salvi and Matteo Pagliardini and Simin Fan and Andreas Köpf and Amirkeivan Mohtashami and Alexandre Sallinen and Alireza Sakhaeirad and Vinitra Swamy and Igor Krawczuk and Deniz Bayazit and Axel Marmet and Syrielle Montariol and Mary-Anne Hartley and Martin Jaggi and Antoine Bosselut},
      year={2023},
      eprint={2311.16079},
      archivePrefix={arXiv},
      primaryClass={cs.CL},
      url={https://arxiv.org/abs/2311.16079}, 
}


\section{Appendix}

This appendix describes the integration of Flash Attention 2 with sequence packing to improve the efficiency of continual pre-training. We outline the main characteristics of Flash Attention 2, explain how sequence packing reduces computational waste compared to traditional padding, and describe the attention masking mechanism that ensures proper isolation between packed sequences.

\subsection{Flash Attention 2: Technical Overview}

Flash Attention 2 builds upon the original Flash Attention algorithm \cite{DBLP:conf/nips/DaoFERR22}, introducing several optimizations that reduce memory traffic between GPU hierarchies while preserving the exact results of standard attention \cite{DBLP:conf/iclr/Dao24}.

The algorithm relies on three main techniques:
\textbf{(i) Tiling}, which partitions computations into blocks that fit in the GPU’s on-chip memory (SRAM), minimizing global memory access;
\textbf{(ii) Recomputation}, which discards intermediate results during the forward pass and recomputes them in the backward pass to save memory; and
\textbf{(iii) Kernel fusion}, which merges multiple GPU operations into single kernels to reduce launch overhead.

These optimizations result in faster trainings and lower memory consumption compared to standard implementations, while maintaining numerical stability and convergence behavior.

\subsection{Sequence packing vs. Traditional padding}

In conventional training pipelines, sequences within the same mini-batch are padded to a uniform length, forcing the model to process large numbers of padding tokens, which carry no information. Sequence packing mitigates this inefficiency by concatenating multiple variable-length sequences into a single tensor while storing the information about their boundaries.

 This method eliminates the computational cost of padding tokens, increases the ratio of meaningful tokens per batch, and improves GPU memory utilization. It was particularly helpful for our Italian medical datasets, where even small efficiency gains have a large cumulative impact, since many trainings were performed to find the best configuration of hyperparameters. This enables efficient processing of sequences up to 1024 tokens on three Nvidia L40S GPUs, maintaining sample independence throughout training.

\subsection{Data collators and Attention masking}

We used the \textbf{PaddingFreeCollator} \cite{kundu2024enhancingtrainingefficiencyusing}, which generalizes the concept of sequence packing through a dynamic batching strategy.

Flash Attention 2 supports 2D attention masks that define explicit boundaries between packed sequences. These masks define precise attention boundaries, ensuring that tokens can only attend to others within the same original sequence. This prevents any interaction between different samples in the packed tensor, while maintaining proper causal masking for autoregressive training.

\begin{figure}[h]
    \includegraphics[scale=0.28]{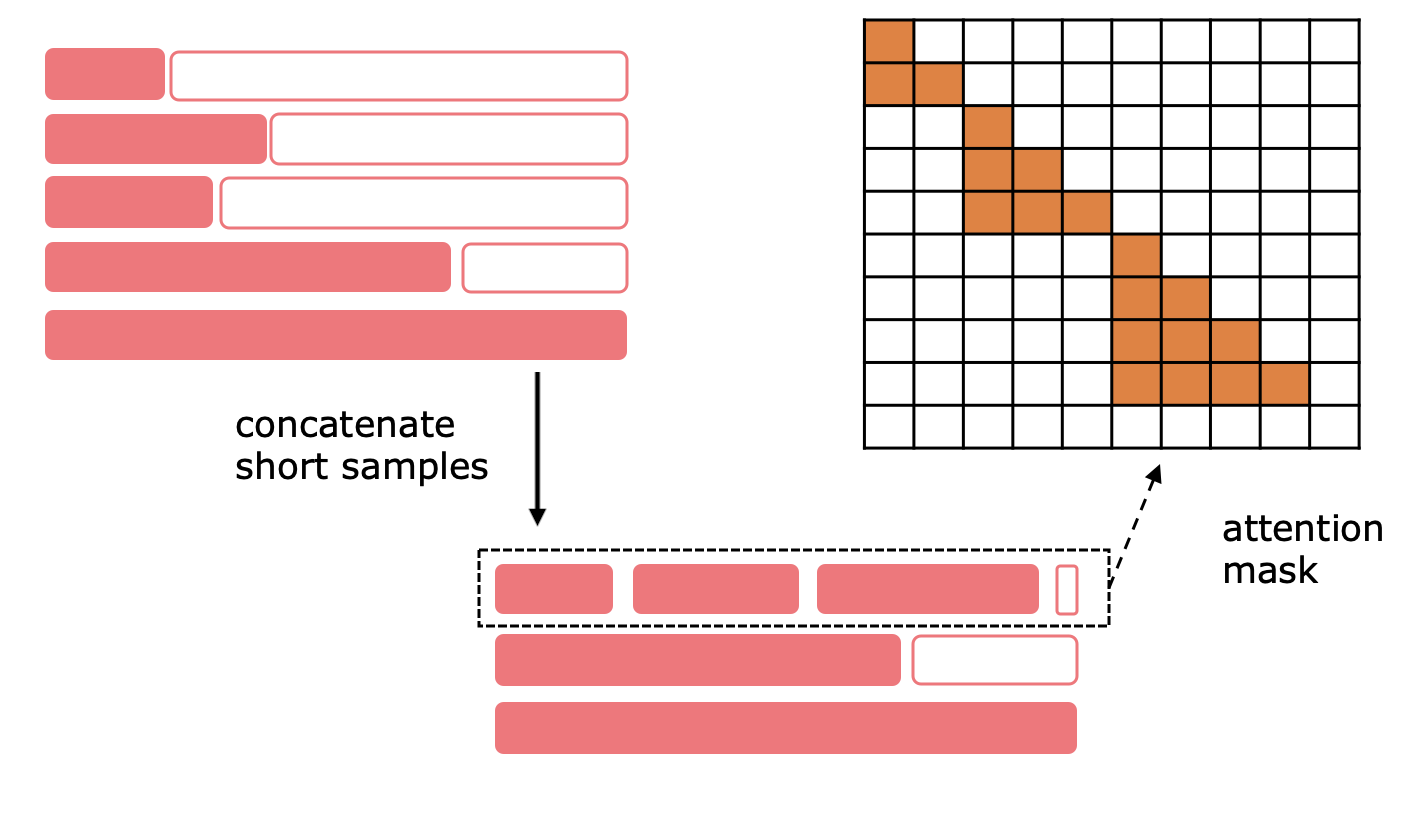}
    \caption{Visualization of sequence packing and the corresponding 2D attention mask that prevents cross-sample attention.}
    \label{fig:sequecepacking}
\end{figure}

\subsection{Implementation and Results}

Flash Attention 2 was enabled in the \textit{transformers} library by setting \texttt{attn\_implementation} to \texttt{flash\_attention\_2} on model instantiation. Combined with the PaddingFreeCollator, this setup allowed for efficient concatenation of sequences while enforcing correct attention masking.

The resulting configuration allows continual pre-training runs to require 8 to 12 hours depending on model size, demonstrating the scalability and efficiency of this approach for domain adaptation tasks even on our reduced setup.

\section{Appendix}

\begin{table*}[!h]
\centering
\begin{tabular}{l|rr|rr|r}
\toprule
 &  \multicolumn{2}{c}{\textbf{4-shot vs 0-shot}} & \multicolumn{2}{|c}{\textbf{CD vs w/o CD}} &  \multicolumn{1}{|c}{\textbf{AVG}} \\
 \midrule
                  & w/o CD & CD & 0-shot & 4-shot & all\\
                  \midrule
gemma-3-1b-pt   & 9.1                  & 8.7              & 4.4              & \underline{4.0}        &     6.6 \\
gemma-3-1b-it    & 4.1                  & 8.2              & 1.7              & \textbf{5.8}       & 5.0      \\
Llama-3.2-1B        & \underline{14.8}                 & \textbf{12.3}             & \underline{5.4}              & 2.9     & 8.9         \\
Llama-3.2-1B-Instruct     & 0.0                  & -1.9             & 2.2              & 0.3    &  0.2          \\
Qwen3-1.7B-Base          & \textbf{25.0}                 & 11.4             & \textbf{14.1}             & 0.5        &   12.8    \\
Qwen3-1.7B         & 9.6                  & \underline{11.6 }            & 1.0              & 3.0       &   6.3     \\
\midrule
\textbf{AVG}        & 10.4                 & 8.4              & 4.8              & 2.8       &       \\
                  \bottomrule
\end{tabular}
\caption{Analysis of the impact of different methods to enhance performances at inference time. The first two columns (\textit{4-shot vs 0-shot}) report the increase of performances moving from 0- to 4-shot with and without Constraint Decoding, while the other two (\textit{CD vs w/o CD}) the increase of performances moving from not using Constraint Decoding to using it in both 0- and 4- shot.}
\label{tab:zoom_inf}
\end{table*}

\begin{table*}[!h]
\centering
\begin{tabular}{l|rrrr}
\toprule
\textbf{Model} & \textbf{0-shot} & \textbf{0-shot + CD} & \textbf{4-shot} & \textbf{4-shot + CD} \\
\midrule
gemma-3-1b-pt             & 385 & 360 & 488 & 403\\
gemma-3-1b-it             & 194 & 202 & 227 & 234\\
Llama-3.2-1B          & 113 & 131 & 135 & 151\\
Llama-3.2-1B-Instruct & 91  & 103 & 136 & 152\\
Qwen3-1.7B                  & 64  & 71  & 173 & 180\\
\bottomrule
\end{tabular}
\caption{Inference time (in seconds) across different configurations for five selected models. Reported time refers to the execution of inference over the 14 sub-tasks on a single H100, serving them through the \textit{vLLM} library with \textit{outlines} for constraint decoding. Overall, inference results in around one hour. A notable difference can be observed among models, some being much faster than others. This is primarily due to the different average output length, as well as the differences in structures.}
\label{tab:inf_time}
\end{table*}

\begin{table*}[]
\centering
\small
\begin{tabular}{ll|rrrrrrr}
\toprule

&& \multicolumn{7}{c}{\textbf{Named Entity Recognition}}  \\
\textbf{Model} & \textbf{Method} & \textbf{cardioccc} & \multicolumn{2}{c}{\textbf{e3c}} & \multicolumn{4}{c}{\textbf{pharmaer}} \\
 &  & \small\textit{medication} & \small\textit{bodypart} & \small\textit{clinical} & \small\textit{anat. part} & \small\textit{disease} & \small\textit{drugs} & \small\textit{symptoms} \small\\
\midrule

\textbf{gemma-3-1b} & 0-shot & 0 & 0 & 0 & 0 & 0 & 0 & 0 \\
\textbf{base} & + CD & 12.98 & 0.95 & 9.99 & 0 & 0 & 9.92 & 0 \\
 & + 4-shot & 5.78 & 0 & 0 & 0 & 2.67 & 15.65 & 3.51 \\
 & + CD,  4-shot & 26.01 & 0 & 4.42 & 0 & 10.2 & 32.86 & 14.93 \\
 & + FT & 99.9 & 32.64 & 57.55 & 67.35 & 48 & 63.93 & 51.55 \\
 & + CPT, FT & 70.46 & 27.72 & 48.99 & 52.38 & 35.4 & 60 & 41.38 \\

\midrule
\textbf{gemma-3-1b} & 0-shot & 32.5 & 2.49 & 20.08 & 14.14 & 13.66 & 41.96 & 23.16 \\
\textbf{instruct} & + CD & 27.25 & 2.55 & 18.38 & 15.83 & 16.38 & 38.26 & 21.67 \\
 & + 4-shot & 41.68 & 4.63 & 15.63 & 3.39 & 10.75 & 29.49 & 12.7 \\
 & + CD,  4-shot & 41.96 & 4.51 & 15.85 & 3.39 & 10.64 & 32.05 & 9.68 \\
 & + FT & 99.95 & 33.22 & 58.59 & 72 & 48.85 & 65.14 & 52.94 \\
 & + CPT, FT & 99.22 & 31.86 & 24.08 & 54.29 & 56.68 & 50.49 & 58.37 \\

\midrule
\textbf{Llama-3.2-1B} & 0-shot & 0 & 0 & 0 & 0 & 0 & 0 & 0 \\
\textbf{base} & + CD & 11.13 & 0 & 0.43 & 0 & 0 & 17.91 & 17.95 \\
 & + 4-shot & 32.52 & 0 & 6.9 & 0 & 9.52 & 31.58 & 3.28 \\
 & + CD,  4-shot & 33.96 & 0.86 & 9.61 & 0 & 8.89 & 32.84 & 17.91 \\
 & + FT & 19.29 & 9.28 & 28.14 & 62.5 & 46.43 & 61.54 & 46.15 \\
 & + CPT, FT & 97.79 & 7.61 & 34.74 & 49.35 & 48.98 & 58.93 & 37.21 \\

\midrule
\textbf{Llama-3.2-1B} & 0-shot & 33.39 & 1.43 & 14.01 & 11.94 & 3.41 & 29.15 & 17.19 \\
\textbf{instruct} & + CD & 33.68 & 1.55 & 13.07 & 9 & 8.96 & 36.67 & 17.48 \\
 & + 4-shot & 29.96 & 3.72 & 14.42 & 0 & 7.79 & 16.3 & 0 \\
 & + CD,  4-shot & 29.05 & 3.93 & 15.11 & 3.7 & 10.26 & 16.42 & 0 \\
 & + FT & 97.95 & 22.14 & 56.01 & 64.71 & 52.94 & 56.76 & 46.43 \\
 & + CPT, FT & 99.95 & 28.31 & 58.45 & 66 & 54.26 & 59.23 & 51.92 \\

\midrule
\textbf{Qwen3-1.7B} & 0-shot & 0 & 0 & 0 & 0 & 0 & 0 & 0 \\
\textbf{base} & + CD & 51 & 0 & 0 & 39.2 & 72.8 & 21.57 & 15.79 \\
 & + 4-shot & 54 & 26.29 & 17.45 & 35.2 & 59.2 & 6.45 & 16.22 \\
 & + CD,  4-shot & 54.2 & 26.83 & 26.11 & 37.6 & 75.2 & 8.82 & 14.95 \\
 & + FT & 99.95 & 38.19 & 59.07 & 72.07 & 55.88 & 61.95 & 48.15 \\
 & + CPT, FT & 99.52 & 23.89 & 57.18 & 72.38 & 60 & 53.39 & 49.48 \\

\midrule
\textbf{Qwen3-1.7B} & 0-shot & 55.51 & 1.49 & 29.02 & 3.85 & 0 & 28.99 & 6.9 \\
\textbf{instruct} & + CD & 55.12 & 1.49 & 28.96 & 3.85 & 0 & 32.65 & 6.9 \\
 & + 4-shot & 59.73 & 3.33 & 29.01 & 13.33 & 3.03 & 35.76 & 0 \\
 & + CD,  4-shot & 59.25 & 3.33 & 29.07 & 13.33 & 3.03 & 34.62 & 0 \\
 & + FT & 100 & 34 & 59.78 & 81.9 & 57.55 & 55.42 & 49.6 \\
 & + CPT, FT & 100 & 38.13 & 60.26 & 69.9 & 57.55 & 56.54 & 48 \\

\bottomrule
\end{tabular}
\caption{Results (F1 score) per subtask in Named Entity Recognition (NER)}
\label{tab:results_ner}
\end{table*}

\begin{table*}[]
\centering
\small
\begin{tabular}{ll|rrrrrrr}
\toprule

&& \multicolumn{7}{c}{\textbf{Question Answering, Relation Extraction, CRF, Argument Mining}}  \\
\midrule
\textbf{Model} & \textbf{Method} 
& \multicolumn{2}{c}{\textbf{QA}} 
& \textbf{RE} 
& \multicolumn{3}{c}{\textbf{CRF}} 
& \textbf{ARG} \\
 &  
& \small\textit{basic} 
& \small\textit{expert ctx} 
& \small\textit{pertainsto} 
& \small\textit{diagnosis} 
& \small\textit{history} 
& \small\textit{rml} 
& \small\textit{arglong} \\
\midrule

\textbf{gemma-3-1b} & 0-shot & 0 & 0 & 0 & 0 & 0 & 0 & 0 \\
\textbf{base} & + CD & 16 & 17.6 & 0 & 0 & 0.44 & 0.56 & 0 \\
 & + 4-shot & 25.6 & 20.8 & 3.12 & 14.09 & 18.21 & 7.43 & 1.91 \\
 & +CD,  4-shot & 25.6 & 20.8 & 4.11 & 14.17 & 18.26 & 7.43 & 12.41 \\
 & + FT & 39.2 & 65.6 & 23.42 & 80.95 & 66.67 & 53.59 & 76.5 \\
 & + CPT, FT & 16.8 & 21.6 & 5.56 & 65.82 & 44.64 & 17.09 & 57.51 \\

\midrule
\textbf{gemma-3-1b} & 0-shot & 28 & 47.2 & 0.13 & 9.24 & 13.57 & 2 & 4.25 \\
\textbf{instruct} & + CD & 20 & 44.8 & 0 & 67.53 & 11.02 & 0.63 & 1.25 \\
 & + 4-shot & 27.2 & 36.8 & 0 & 31.96 & 25.56 & 17.73 & 17.8 \\
 & +CD,  4-shot & 28.8 & 45.6 & 0.57 & 65.57 & 24.03 & 17.92 & 30.5 \\
 & + FT & 37.6 & 65.6 & 24 & 81.93 & 67.43 & 56.25 & 78.5 \\
 & + CPT, FT & 52 & 71.7 & 47.2 & 82.76 & 70.83 & 43.75 & 68.75 \\

\midrule
\textbf{Llama-3.2-1B} & 0-shot & 0 & 0 & 0 & 0 & 0 & 0 & 0 \\
\textbf{base} & + CD & 12.8 & 18.4 & 0 & 9.47 & 0.53 & 0 & 1.15 \\
 & + 4-shot & 24 & 27.2 & 3.15 & 14.25 & 15.75 & 3.39 & 22.04 \\
 & +CD,  4-shot & 24.8 & 27.2 & 3.08 & 20.07 & 13.61 & 5.88 & 31.58 \\
 & + FT & 47.2 & 68 & 30.56 & 41.03 & 65.57 & 47.87 & 70.56 \\
 & + CPT, FT & 40 & 35.2 & 20.81 & 10.37 & 78.01 & 15.3 & 74.78 \\

\midrule
\textbf{Llama-3.2-1B} & 0-shot & 31.2 & 54.4 & 1.31 & 4.66 & 48.24 & 1.33 & 7.19 \\
\textbf{instruct} & + CD & 32.8 & 53.6 & 1.25 & 35.71 & 46.59 & 1.34 & 6.63 \\
 & + 4-shot & 31.2 & 40.8 & 6.27 & 39.25 & 26.97 & 4.55 & 8.68 \\
 & +CD,  4-shot & 30.4 & 40.8 & 6.24 & 0 & 26.97 & 4.57 & 23.09 \\
 & + FT & 49.6 & 68.8 & 30.88 & 85.71 & 68.39 & 60.98 & 76.58 \\
 & + CPT, FT & 37.6 & 48.8 & 27.45 & 80.9 & 65.91 & 47.41 & 73.79 \\

\midrule
\textbf{Qwen3-1.7B} & 0-shot & 0 & 0 & 0 & 0 & 0 & 0 & 0 \\
\textbf{base} & + CD & 15.8 & 0 & 1.04 & 8.03 & 13.3 & 0 & 25.75 \\
 & + 4-shot & 16.99 & 7.54 & 2.12 & 56.06 & 15.57 & 17.01 & 50.24 \\
 & +CD,  4-shot & 18.69 & 7.88 & 2.61 & 56.84 & 15.95 & 0 & 52.74 \\
 & + FT & 41.6 & 80 & 33.17 & 87.06 & 73.49 & 60.61 & 79.39 \\
 & + CPT, FT & 38.4 & 67.2 & 18.28 & 80.49 & 69.84 & 60.61 & 73.98 \\

\midrule
\textbf{Qwen3-1.7B} & 0-shot & 36 & 76 & 4.09 & 60.34 & 35.1 & 0 & 11.16 \\
\textbf{instruct} & + CD & 38.4 & 75.2 & 3.98 & 70.71 & 35.1 & 0 & 11.36 \\
 & + 4-shot & 33.6 & 55.2 & 12.27 & 60.74 & 39.71 & 10.42 & 54.52 \\
 & +CD,  4-shot & 39.2 & 79.2 & 12.4 & 61.19 & 39.61 & 10.42 & 54.91 \\
 & + FT & 47.2 & 80.8 & 37.86 & 86.67 & 69.32 & 62.58 & 81.97 \\
 & + CPT, FT & 44.8 & 70.4 & 25.33 & 82.93 & 71.43 & 47.74 & 77.61 \\

\bottomrule
\end{tabular}
\caption{Results (F1 score for RE, CRF, ARG; accuracy for QA) per subtask for Queston Answering (QA), Relation Extraction (RE), Case Report Form filling (CRF) and Argument Mining (ARG).}
\label{tab:results_other_tasks}
\end{table*}

\begin{table*}[]
\centering
\small
\begin{tabular}{ll|rrrrrr}
\toprule

&& \multicolumn{6}{c}{\textbf{Out Of Distribution}} \\
\midrule
\textbf{Model} & \textbf{Method}
& \multicolumn{3}{c}{\textbf{QA}}
& \multicolumn{3}{c}{\textbf{NER}} \\
 & 
& \small\textit{at 2023}
& \small\textit{medmcqa}
& \small\textit{medqa}
& \small\textit{e3c clinical}
& \small\textit{distemist}
& \small\textit{psynit} \\
\midrule

\textbf{gemma-3-1b} & 0-shot & 28.0 & 27.5 & 28.1 & 25.5 & 21.5 & 17.9 \\
\textbf{instruct} & + CD & 26.8 & 30.2 & 29.9 & 22.9 & 21.1 & 22.0 \\
 & + 4-shot & 27.0 & 27.2 & 30.8 & 27.0 & 23.1 & 25.1 \\
 & +CD,  4-shot & 28.4 & 28.5 & 31.5 & 27.4 & 25.7 & 23.8 \\
 & + CPT, FT & 28.4 & 27.1 & 31.8 & 54.5 & 37.5 & 27.2 \\

\midrule
\textbf{Llama-3.2-1B} & 0-shot & 29.0 & 29.7 & 31.6 & 17.3 & 25.7 & 12.1 \\
\textbf{instruct} & + CD & 27.6 & 29.4 & 31.5 & 16.7 & 25.4 & 11.9 \\
 & + 4-shot & 28.6 & 29.1 & 30.1 & 23.4 & 16.2 & 13.7 \\
 & +CD,  4-shot & 29.0 & 29.1 & 30.6 & 25.3 & 16.6 & 13.0 \\
 & + FT & 31.0 & 27.4 & 30.2 & 51.4 & 43.6 & 31.9 \\

\midrule
\textbf{Qwen3-1.7B} & 0-shot & 0.2 & 0.4 & 0.5 & 2.3 & 22.8 & 25.8 \\
\textbf{base} & + CD & 48.6 & 36.6 & 35.9 & 32.1 & 19.1 & 25.8 \\
 & + 4-shot & 39.0 & 29.2 & 35.4 & 31.0 & 24.8 & 27.1  \\
 & +CD,  4-shot & 51.8 & 37.7 & 36.1 & 32.8 & 15.7 & 25.3 \\
 & + FT & 39.4 & 28.6 & 33.9 & 53.8 & 44.7 & 35.2 \\

\bottomrule
\end{tabular}
\caption{Results (F1 score for NER; accuracy for QA) per subtask for Question Answering (QA) and Named Entity Recognition (NER).}
\label{tab:results_ood_all}
\end{table*}

\begin{table*}[]
\centering
\begin{tabular}{cl|rrr}
\toprule
\textbf{Model} & \textbf{Method} & \multicolumn{3}{|c}{\textbf{AVG F1} ($\delta$ baseline)} \\
& &\textit{sample-weight}& \textit{per sub-task }& \textit{per task} \\
\midrule
 \textbf{Qwen3-32B}  & \textit{+ 4-shot} & 52.0&56.8& 54.7 \\
 \midrule
\midrule
 & \textit{0-shot}             & 0.0 \small\textcolor{red}{-52.0}   &  0.0    \small\textcolor{red}{-56.8}   &       0.0 \small\textcolor{red}{-54.7}  \\
 & \textit{+ CD}                 & 2.8 \small\textcolor{red}{-49.2}   &  4.9  \small\textcolor{red}{-51.9}      &4.4   \small\textcolor{red}{-50.4}  \\
 & \textit{+ 4-shot}        & 8.4 \small\textcolor{red}{-43.5}   &   8.5      \small\textcolor{red}{-48.3} &9.1    \small\textcolor{red}{-45.7}  \\
 & \textit{+ CD, 4-shot}  & 10.4 \small\textcolor{red}{-41.6}   &  13.7     \small\textcolor{red}{-43.2} &13.1  \small\textcolor{red}{-41.6}  \\
 & \textit{+ FT} & \underline{56.5}  \small\textcolor{ForestGreen}{+4.6} &\underline{59.1} \small\textcolor{ForestGreen}{+2.3} &55.9  \small\textcolor{ForestGreen}{+1.2}  \\
\multirow{-6}{*}{\makecell[c]{\textbf{gemma-3-1b} \\\textbf{base}\\ \small 1.00B params}} & \textit{+ CPT, FT} & 37.3 \small\textcolor{red}{-14.6} &40.4 \small\textcolor{red}{-16.4} &34.6  \small\textcolor{red}{-20.2}   \\

\midrule
 & \textit{0-shot}& 10.8        \small\textcolor{red}{-41.2} &  18.0\small\textcolor{red}{-38.8} &14.3   \small\textcolor{red}{-40.5}  \\
 & \textit{+ CD}  & 18.9        \small\textcolor{red}{-33.0} & 20.4 \small\textcolor{red}{-36.4} &16.0   \small\textcolor{red}{-38.7}  \\
 & \textit{+ 4-shot} & 18.6     \small\textcolor{red}{-33.4} & 19.7 \small\textcolor{red}{-37.1} &18.4    \small\textcolor{red}{-36.4}  \\
 & \textit{+ CD, 4-shot}  & 24.5\small\textcolor{red}{-27.5} & 23.6 \small\textcolor{red}{-33.2} &24.2    \small\textcolor{red}{-30.5}  \\
 & \textit{+ FT} &  \textbf{57.6} \small\textcolor{ForestGreen}{+5.7} &\textbf{60.1}  \small\textcolor{ForestGreen}{+3.3}  &\underline{56.8}  \small\textcolor{ForestGreen}{+2.1}  \\
\multirow{-6}{*}{\makecell[c]{\textbf{gemma-3-1b}\\ \textbf{instruct}\\ \small 1.00B params}} & \textit{+ CPT, FT} & 54.7 \small\textcolor{ForestGreen}{+2.8} &58.0 \small\textcolor{ForestGreen}{+1.2} &\textbf{59.4}   \small\textcolor{ForestGreen}{+4.7}  \\
\midrule
\midrule

 & \textit{0-shot} &    0.0    \small\textcolor{red}{-56.8} &0.0  \small\textcolor{red}{-56.8} &0.0   \small\textcolor{red}{-54.7}  \\
 & \textit{+ CD} &       3.2   \small\textcolor{red}{-50.4} &6.4  \small\textcolor{red}{-50.4} &5.4    \small\textcolor{red}{-49.4}  \\
 & \textit{+ 4-shot}& 9.9      \small\textcolor{red}{-43.0} &13.8 \small\textcolor{red}{-43.0} &14.8  \small\textcolor{red}{-40.0}  \\
 & \textit{+ CD, 4-shot}& 11.8 \small\textcolor{red}{-40.4} &15.0 \small\textcolor{red}{-40.4} &17.7  \small\textcolor{red}{-37.0}  \\
 & \textit{+ FT} &        41.0 \small\textcolor{red}{-10.9} &46.0 \small\textcolor{red}{-10.8} &49.9  \small\textcolor{red}{-4.9}  \\
\multirow{-6}{*}{\makecell[c]{\textbf{Llama-3.2-1B} \\\textbf{base}\\ \small 1.24B params}} & \textit{+ CPT, FT} & 33.3  \small\textcolor{red}{-18.7} &43.5  \small\textcolor{red}{-13.3} &43.1 \small\textcolor{red}{-11.6}  \\
\midrule
 & \textit{0-shot}& 14.8      \small\textcolor{red}{-37.2} &18.5 \small\textcolor{red}{-38.3} &17.0 \small\textcolor{red}{-37.7}  \\
 & \textit{+ CD} &  19.6       \small\textcolor{red}{-32.3} &21.3 \small\textcolor{red}{-35.5} &19.2 \small\textcolor{red}{-35.5}  \\
 & \textit{+ 4-shot} & 14.8     \small\textcolor{red}{-34.4} &18.5 \small\textcolor{red}{-40.4} &17.0 \small\textcolor{red}{-37.8}  \\
 & \textit{+ CD, 4-shot} & 11.7 \small\textcolor{red}{-40.3} &15.0 \small\textcolor{red}{-41.8} &17.3 \small\textcolor{red}{-37.4}  \\
 & \textit{+ FT}  & \textbf{58.4} \small\textcolor{ForestGreen}{+6.4} &\textbf{59.8}  \small\textcolor{ForestGreen}{+3.0} &\textbf{59.0} \small\textcolor{ForestGreen}{+4.3}  \\
\multirow{-6}{*}{\makecell[c]{\textbf{Llama-3.2-1B} \\\textbf{instruct}\\ \small 1.24B params}} & \textit{+ CPT, FT} & \underline{55.0}  \small\textcolor{ForestGreen}{+3.0}  &\underline{57.1} \small\textcolor{ForestGreen}{+0.3} &\underline{53.8}  \small\textcolor{red}{-1.0}  \\
\midrule
\midrule

 & \textit{0-shot}& 0.0          \small\textcolor{red}{-52.0} & 0.0   \small\textcolor{red}{-56.8} & 0.0 \small\textcolor{red}{-54.7}  \\
 & \textit{+ CD}& 7.9            \small\textcolor{red}{-44.1} &18.9   \small\textcolor{red}{-37.9} &14.1 \small\textcolor{red}{-40.7}  \\
 & \textit{+ 4-shot} & 24.8      \small\textcolor{red}{-27.2} &27.2   \small\textcolor{red}{-29.6} &25.0 \small\textcolor{red}{-29.8}  \\
 & \textit{+ CD, 4-shot} & 23.7  \small\textcolor{red}{-28.3} &28.5   \small\textcolor{red}{-28.4} &25.5 \small\textcolor{red}{-29.2}  \\
 & \textit{+ FT}  & \underline{62.4} \small\textcolor{ForestGreen}{+10.4}  &\underline{63.6} \small\textcolor{ForestGreen}{+6.8} &\underline{61.9}   \small\textcolor{ForestGreen}{+7.1}  \\
\multirow{-6}{*}{\makecell[c]{\textbf{Qwen3-1.7B} \\\textbf{base}\\ \small 1.72B params}} & \textit{+ CPT, FT} & 56.4 \small\textcolor{ForestGreen}{+4.4}  &58.9 \small\textcolor{ForestGreen}{+2.1} &55.0  \small\textcolor{ForestGreen}{+0.2}  \\
\midrule
 & \textit{0-shot}  & 24.8      \small\textcolor{red}{-27.2} & 24.9 \small\textcolor{red}{-31.9} &24.2  \small\textcolor{red}{-30.5}  \\
 & \textit{+ CD} &    26.5      \small\textcolor{red}{-25.5} & 26.0 \small\textcolor{red}{-30.8} &25.2  \small\textcolor{red}{-29.6}  \\
 & \textit{+ 4-shot}  & 29.2    \small\textcolor{red}{-22.7} & 29.3 \small\textcolor{red}{-27.5} &33.8  \small\textcolor{red}{-21.0}  \\
 & \textit{+ CD, 4-shot}  & 20.0\small\textcolor{red}{-22.0} & 31.4 \small\textcolor{red}{-25.4} &36.8  \small\textcolor{red}{-17.9}  \\
 & \textit{+ FT} &  \textbf{62.4}  \small\textcolor{ForestGreen}{+10.4}  &\textbf{64.6}  \small\textcolor{ForestGreen}{+7.8} &\textbf{63.9}  \small\textcolor{ForestGreen}{+9.1}  \\
\multirow{-6}{*}{\makecell[c]{\textbf{Qwen3-1.7B} \\\textbf{instruct}\\ \small 1.72B params}} & \textit{+ CPT, FT} & 58.2 \small\textcolor{ForestGreen}{+6.2} &60.8 \small\textcolor{ForestGreen}{+3.9}  &57.9 \small\textcolor{ForestGreen}{+3.1}  \\
\bottomrule
\end{tabular}     
\caption{Average performances of models and adaptation techniques by aggregation strategy. The five tasks are composed by $14$ subtasks, unevenly distridubets. Each subtask is the combination of a task and a dataset. Each dataset has a different number of testing examples. We report the F1 score averaged by number of examples in each task (\textit{sample-weight}), by number of sub-tasks per task, and by tasks.
For each model family, the best and second-best performances are in \textbf{bold} and \underline{underlined} respectively.
It can be noticed that there is not significant difference among aggregation method.}
\label{tab:small_models_results_avg_methods}
\end{table*}


\end{document}